\pdfoutput=1
\documentclass{article}
\usepackage{microtype}
\usepackage{graphicx}
\usepackage{subcaption}
\usepackage{booktabs}
\usepackage{hyperref}

\usepackage[accepted]{icml2026}
\usepackage{amsmath}
\usepackage{amssymb}
\usepackage{mathtools}
\usepackage{amsthm}
\usepackage{dblfloatfix}
\usepackage{ml_defs}
\usepackage{xspace}
\usepackage[capitalize,noabbrev]{cleveref}
\theoremstyle{plain}

\theoremstyle{definition}

\theoremstyle{remark}

\usepackage[textsize=tiny]{todonotes}

\newcommand{\ourMethod}{\textsc{WIND}\xspace}

\renewcommand{\paragraph}[1]{\textbf{#1}. }
\icmltitlerunning{WIND}

\begin{document}

\twocolumn[

 \icmltitle{WIND: Weather Inverse Diffusion for Zero-Shot Atmospheric Modeling}
  \icmlsetsymbol{equal}{*}
  \begin{icmlauthorlist}
    \icmlauthor{Michael Aich}{equal,tum,potsdam}
    \icmlauthor{Andreas Fürst}{equal,jku}
    \icmlauthor{Florian Sestak}{equal,jku,emmi}
    \icmlauthor{Carlos Ruiz-Gonzalez}{jku}
    \icmlauthor{Niklas Boers}{tum,potsdam,exeter}
    \icmlauthor{Johannes Brandstetter}{jku,emmi}
  \end{icmlauthorlist}

  \icmlaffiliation{tum}{Munich Climate Center and Earth System Modelling Group, TUM School of Engineering and Design, Technical University of Munich, Germany}
  \icmlaffiliation{jku}{ELLIS Unit, LIT AI Lab, Institute for Machine Learning, JKU Linz, Austria}
  \icmlaffiliation{emmi}{Emmi AI GmbH, Linz, Austria}
  \icmlaffiliation{potsdam}{Potsdam Institute for Climate Impact Research, Potsdam, Germany}
  \icmlaffiliation{exeter}{Department of Mathematics, University of Exeter, Exeter, United Kingdom}
  \icmlcorrespondingauthor{Michael Aich}{michael.aich@tum.de}
  \icmlcorrespondingauthor{Andreas Fürst}{fuerst@ml.jku.at}
  \icmlkeywords{Machine Learning, ICML}

  \vskip 0.3in
]

\printAffiliationsAndNotice{\icmlEqualContribution}

\begin{abstract}
    Deep learning has revolutionized weather forecasting, but many challenges remain, including climate modeling. Moreover, the current landscape remains fragmented: highly specialized models are typically trained individually for distinct tasks. To unify this landscape, we introduce WIND, a single pre-trained foundation model capable of replacing specialized baselines across a vast array of tasks. Crucially, in contrast to previous atmospheric foundation models, we achieve this without any task-specific fine-tuning. 
    To learn a robust, task-agnostic prior of the atmosphere, we pre-train WIND with a self-supervised video reconstruction objective, utilizing an unconditional video diffusion model to iteratively reconstruct atmospheric dynamics from a noisy state. At inference, we frame diverse domain-specific problems strictly as inverse problems and solve them via posterior sampling.
    This unified approach allows us to tackle highly relevant weather and climate problems, including probabilistic forecasting, spatial and temporal downscaling, reconstruction of spatial fields from sparse observations and enforcing global dry air mass conservation. We further demonstrate how WIND can be applied to explore extreme weather events under prescribed out-of-distribution thermodynamic perturbations. By combining generative video modeling with inverse problem solving, WIND offers a computationally efficient alternative for AI-based atmospheric modeling.
\end{abstract}

\section{Introduction}

Understanding atmospheric dynamics under climate change is of utmost importance. Adverse atmospheric conditions drive severe humanitarian and financial crises, with global economic costs exceeding $\$$4.3 trillion over the past 50 years \citep{SwissRe2024, WMO2023}. In terms of societal impact, precipitation is among the most critical variables. Extreme precipitation events lead to devastating floods and landslides, whose frequency and intensity increase with global warming \citep{IPCC_AR6_SYR_SPM_2023}. Beyond disaster risk mitigation, atmospheric modeling is central to the energy transition. Wind speed, for instance, is a key predictor of power output and economic viability for renewable energy projects \citep{yan2025economic}.

Petabyte-scale atmospheric datasets, like ERA5 \citep{hersbach2020era5}, which offer decades of high-resolution data, are both a challenge and an opportunity for weather forecasting. While the predictive skill of classical numerical weather prediction (NWP) models scales primarily with increased computational power rather than historical dataset size \citep{bauer2015quiet}, this data abundance directly fuels atmospheric foundation models \citep{bodnar2025foundation, nguyen2023climax, lessig2023atmorep}. Pre-trained on massive datasets, these models capture the underlying dynamics and relationships of the atmosphere, encoding them into compact representations.

Foundation models mark a paradigm shift from the current fragmented landscape, where specialized models are trained from scratch for every niche (e.g., precipitation downscaling vs. wind forecasting). Instead, a single foundation model can be pre-trained once and efficiently fine-tuned to diverse downstream applications.
An effective atmospheric foundation model must reflect the chaotic and probabilistic nature of weather data. To address this, we present \ourMethod, a framework that combines diffusion forcing training with moment matching posterior sampling (MMPS) at inference to construct a unified foundation model of the atmosphere. Unlike existing approaches, which train or fine-tune additional networks for each new task \citep{bodnar2025foundation, nguyen2023climax, lessig2023atmorep}, we solve diverse downstream tasks purely at inference time. We use ERA5 data at 1.5° resolution, instead of 0.25°, to mitigate the prohibitive computational training costs. Thus, we focus on demonstrating the conceptual novelty and versatility of our approach, rather than competing directly with state-of-the-art operational baselines.

\textbf{Our contributions.} We propose \ourMethod a probabilistic foundation model of the atmosphere that can solve a large variety of climate and weather specific downstream tasks, eliminating the need for task-specific fine-tuning. \ourMethod can stabilize long rollouts and allows for explicit guidance of the generative process. By explicitly modeling precipitation, we address a critical gap in current baselines.

\paragraph{Conflict of Interest Disclosure}
The authors declare no competing financial interests.

\begin{figure*}[t!]
    \centering
    \includegraphics[width=1 \linewidth]{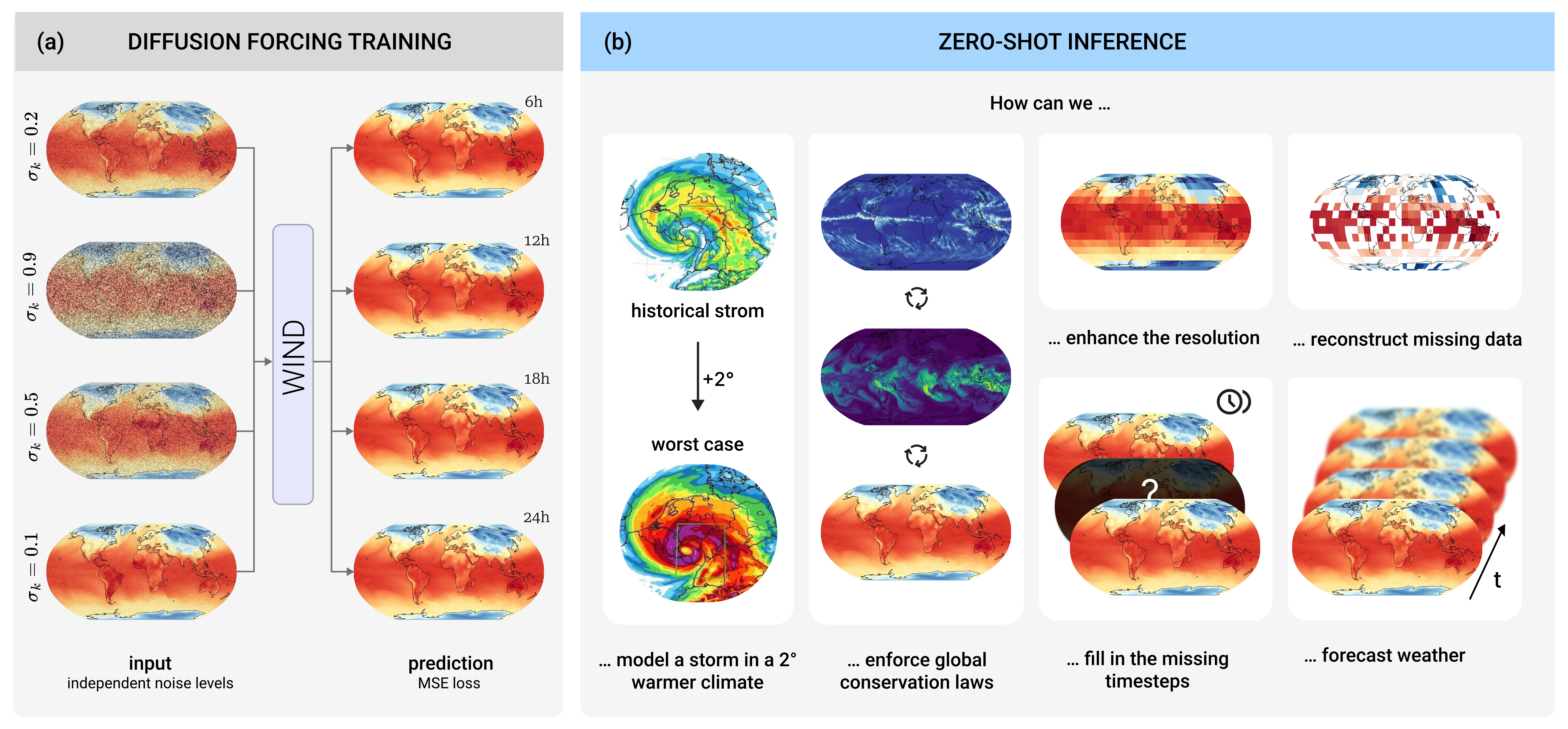}
    \caption{\textbf{Training setup and inference capabilities of \ourMethod} 
    \textbf{(a) Training:} We apply an independent noise level $\sigma_k$ to each frame in the sequence. The model is trained without explicit noise level information. \ourMethod learns to jointly denoise the sequence, enabling it to handle arbitrary combinations of clean and noisy context frames.
    \textbf{(b) Inference:} \ourMethod addresses various climate- and weather-related questions by framing them as inverse problems: recover the atmospheric field $x$ from the observation $y$ that is constrained by some operator $\mathcal{A}$ as $\mathcal{A}(x) = y$. We demonstrate how to formulate a task-specific operator $\mathcal{A}$ for each question in (b).
    }
    \label{fig:methods_fig}
\end{figure*}

\begin{figure*}[t]
    \centering
    \includegraphics[width=1 \linewidth]{figures/figure_2.pdf}
    \caption{
    \textbf{(a) Forecasting with dry air mass guidance:} The model receives one clean frame together with four noisy frames and
    progressively denoises them through repeated DDIM update steps. The procedure is then repeated
    autoregressively for the entire forecast horizon. At every DDIM step,
    guidance is injected using the dry air mass
    $\mathbf{Y}=f_{\mathrm{DAM}}(\mathbf{x}^{0})$ computed from the
    initial clean frame.
    \textbf{(b) Guidance with MMPS:} Within each DDIM update, the noisy
    latent $\mathbf{z}^{t}$ is passed through a data-prediction network to
    obtain the predicted clean state
    $\hat{\mathbf{x}}_{\theta}^{t}$, from which the prior score
    $\mathbf{s}^{t}$ is derived. The forward operator
    $\mathcal{A}$ is applied to both
    $\hat{\mathbf{x}}_{\theta}^{t}$ and the ground-truth state
    $\mathbf{x}^{t}$, yielding the predicted and true measurements
    $\hat{\mathbf{y}}^{t}$ and $\mathbf{y}^{t}$, respectively. MMPS
    compares these to produce the likelihood score
    $\mathbf{s}_{\mathrm{lhd}}^{t}$, which is added to the prior score
    to form the total score used in the DDIM update.
    }
    \label{fig:figure_2}
\end{figure*}

\section{Background and Related Work}

Numerical Weather Prediction (NWP) was long established as the standard for operational weather forecasting, making predictions based on numerical fluid dynamics simulations. However, NWP remains computationally extremely demanding and slow \citep{bauer2015quiet}. Recently data-driven deep learning approaches have emerged as a powerful alternative. By leveraging massive amounts of high-dimensional historical data, these models directly try to learn atmospheric dynamics. They shift the heavy computational burden almost entirely to the training phase, enabling real-time inference that is orders of magnitude faster and more efficient than traditional NWP systems. AI-based forecasting models have rapidly evolved over the last years, closing the performance gap with traditional physical solvers while offering significantly reduced inference costs \citep{price2025probabilistic, AIFS, lam2023learning, bi2023accurate}.

\paragraph{Diffusion models for spatiotemporal data} 
While initial deterministic AI models \citep{bi2023accurate, lam2023learning} dominated the field, they produce blurry predictions at long lead times as a result of minimizing the mean squared error (MSE) during training \citep{price2025probabilistic}. Stochastic diffusion models (see \cref{app:diffusion_background}) solve this by modeling the full probability distribution of high-dimensional data  \citep{ho2020denoising, song2020score}. This allows them to generate sharp, coherent structures that reflect the multi-scale nature of the data. Their inherent stochasticity allows them to generate ensemble forecasts, which is essential for quantifying uncertainty and assessing the probability of extreme weather events \citep{li2023seeds}.
While diffusion models have revolutionized static image synthesis, adapting them to spatiotemporal dynamics requires specific architectural strategies.
As illustrated in \cref{fig:methods_fig}, we frame atmospheric modeling as a video generation task, where atmospheric variables correspond to channels (like RGB in a video) and time steps correspond to video frames. Currently, autoregressive conditional diffusion models, such as GenCast \citep{price2025probabilistic}, represent the state-of-the-art in forecasting. They predict the next state given a context history. However, long autoregressive rollouts suffer from error accumulation, often leading to instabilities or divergence \citep{chen2024diffusion}. Furthermore, they lack a mechanism to guide the entire sequence toward a global objective, as early predictions are fixed and cannot be retroactively adjusted.

\paragraph{Autoregressive vs sequence diffusion}
A common alternative in video generation is full-sequence diffusion \citep{ho2022video}, which diffuses a fixed number of frames simultaneously with a shared noise level. While this enables globally consistent sampling, these models often struggle with generating long rollouts beyond their training window length. Extending the video autoregressively, by using the last clean frame of the previous window as the first frame of the new window, fails because it creates an out-of-distribution state, as the model requires all frames to share the same noise level. Recent efforts like Elucidated Rolling Diffusion Models (ERDM) \citep{cachay2025elucidated, ruhe2024rolling} attempt to mitigate this failure by defining a fixed noise schedule where frames become progressively noisier, effectively encoding the growing uncertainty at longer lead times. However, this rigid structure restricts the model's flexibility during inference.

\paragraph{Unified modeling with diffusion forcing} 
In this work, we leverage diffusion forcing \citep{chen2024diffusion} to overcome the downsides of autoregressive and full-sequence diffusion models. Unlike standard approaches, we train on sequences where each frame is assigned a random, independent noise level. This allows the model to accept clean frames (zero noise) from previous windows as context without distribution shift, enabling stable, arbitrarily long rollouts \citep{chen2024diffusion}. This flexible training objective natively supports rolling diffusion-like inference \citep{ruhe2024rolling, cachay2025elucidated} without requiring rigid, pre-defined denoising-schedules. This strategy aligns with the reconstruction paradigm, one of the two leading frameworks in Self-Supervised Learning \citep{van2025joint}. One can interpret diffusion noise as a continuous masking mechanism: clean frames serve as unmasked context, while noisy frames are partially masked \citep{hu2024mask}. Diffusion forcing thus generalizes the discrete masking objective, enabling the model to learn robust spatiotemporal representations by reconstructing atmospheric dynamics from varying degrees of corruption.

\paragraph{Guidance for diffusion models} 
A critical advantage of diffusion models over autoregressive approaches is the ability to guide the sampling of a full sequence to minimize a global objective. In image generation, guidance is typically achieved via Classifier-Free Guidance (CFG) \citep{ho2022classifier}. 
CFG offers a tradeoff between diversity and sample quality by jointly training an unconditional and a conditional diffusion model. However, extending CFG to temporal sequences is challenging, as it typically requires specific training strategies such as history dropout to learn an unconditional prior. Recent work on history-guided video diffusion \citep{song2025history} overcomes this limitation by leveraging the unique properties of diffusion forcing. Because the model is trained with independent noise levels, it can naturally estimate both conditional (clean history) and unconditional (fully noisy history) scores, enabling CFG at inference time without training a separate classifier. While history guidance targets temporal consistency, our work aims to enforce physical consistency via external constraints. We frame diverse climate and weather tasks as inverse problems, using posterior sampling to guide the generation process. Unlike standard diffusion posterior sampling \citep{chung2022diffusion}, which relies on point estimates, we employ MMPS \citep{rozet2024learning}. 

\section{Approach}\label{sec:approach}

The \ourMethod framework decouples learning of atmospheric dynamics
from specific downstream applications. Our pipeline consists of two stages: (1)~pre-training a spatiotemporal UViT backbone using diffusion forcing with independent per-frame noise levels to learn a flexible generative prior, and (2)~applying Moment Matching Posterior Sampling ~\citep{rozet2024learning} at inference to solve downstream tasks as zero-shot inverse problems.

\paragraph{Problem formulation and notation}
We denote the atmospheric state as a spatiotemporal tensor 
$\mathbf{X} = \{\mathbf{x}^1, \dots, \mathbf{x}^T\} \in \mathbb{R}^{T \times C \times H \times W}$, where $T$ is the sequence length, $C$ the number of variables, and $H, W$ the spatial resolution. We pre-train on sequences of length $T=5$ at a 6-hour stride, covering $C=70$ variables at $1.5^\circ$ resolution ($H=121$, $W=240$) (see \cref{Appendix_dataset}). All variables are processed jointly by the UViT at every denoising step, so inter-variable dependencies and couplings are encoded implicitly in the learned prior $p(\mathbf{Z})$.

\paragraph{Diffusion forcing training}
Standard video diffusion corrupts all frames at the same noise level~\citep{ho2022video}. Following the diffusion forcing paradigm~\citep{chen2024diffusion}, we instead sample a noise level $k^t \in [0,1]$ \emph{independently} for each frame $t \in \{1, \dots, T\}$.
The forward process for frame $\mathbf{x}^t$ is:
\begin{equation}
    \mathbf{z}^t = \alpha(k^t)\mathbf{x}^t + \beta(k^t)\boldsymbol{\epsilon}^t,
    \quad \boldsymbol{\epsilon}^t \sim \mathcal{N}(\mathbf{0}, \mathbf{I}),
\end{equation}
where $\alpha(k^t)$ and $\beta(k^t)$ are the signal and noise
schedule coefficients. The noised sequence is $\mathbf{Z} = \{\mathbf{z}^1, \dots,
\mathbf{z}^T\}$.

The UViT backbone $\hat{\mathbf{X}}_\theta(\mathbf{Z})$ is trained to reconstruct the clean sequence via denoising score matching~\citep{vincent2011connection, hyvarinen2005estimation}. A critical design choice, motivated by \citet{sun2025noise}, is that we do \emph{not} condition the network on the noise levels $k^t$. The model must therefore infer per-frame uncertainty solely from its noisy inputs, preventing over-reliance on explicit schedules and encouraging more robust representations. For training objective and noise schedule details see \cref{app:diffuion_forcing_details}.

\paragraph{Inference via moment matching posterior sampling}
At inference time each downstream task is framed as an inverse problem, to recover the atmospheric state $\mathbf{X}$ from observations $\mathbf{Y}$ related by a forward operator $\mathcal{A}$,
\begin{equation}
    \mathbf{Y} = \mathcal{A}(\mathbf{X}) + \boldsymbol{\eta},
    \quad
    \boldsymbol{\eta} \sim \mathcal{N}(\mathbf{0}, \delta^2\mathbf{I}),
\end{equation}
where $\delta^2$ is a task-specific noise variance.
$\mathcal A$ is specific to each task, often non-linear, and always differentiable. We want to condition the sampling based on the observation $\mathbf Y$, a vector of arbitrary length. The posterior score is given by the Bayes rule,
\begin{equation}\label{eq:bayes}
    \nabla_{\mathbf{Z}} \log p(\mathbf{Z}|\mathbf{Y})
    =\underbrace{\nabla_{\mathbf{Z}} \log p(\mathbf{Z})}_{\text{Prior score}}
    +\underbrace{\nabla_{\mathbf{Z}} \log p(\mathbf{Y}|\mathbf{Z})}_{\text{Likelihood score}},
\end{equation}
where the prior score is provided by our frozen pre-trained model $\mathbf{s}_\theta(\mathbf{Z})$. 
The key challenge lies in estimating the likelihood score
$\nabla_{\mathbf{Z}} \log p(\mathbf{Y}|\mathbf{Z})$, which requires marginalising over all clean atmospheric states:
\begin{equation} \label{eq:p_y_z}
    p(\mathbf{Y}|\mathbf{Z})
    = \int p(\mathbf{Y}|\mathbf{X})\,p(\mathbf{X}|\mathbf{Z})\,
      \mathrm{d}\mathbf{X}.
\end{equation}
Diffusion posterior sampling~\citep{chung2022diffusion} assumes $p(\mathbf{X}|\mathbf{Z})$ as a Dirac delta distribution at the current prediction $\hat{\mathbf{X}}_\theta(\mathbf{Z})$,
\begin{equation}
    p(\mathbf{X}|\mathbf{Z}) \approx \mathcal{N}(\hat{\mathbf{X}}_\theta(\mathbf{Z}), 0 \mathbf I).
\end{equation}
This ignores the model's predictive uncertainty at intermediate noise levels, which can lead to miscalibrated guidance especially in high-noise regimes. 
Moment matching posterior sampling (MMPS) \citep{rozet2024learning} instead considers a Gaussian approximation 
\begin{equation}
    p(\mathbf{X}|\mathbf{Z}) \approx \mathcal{N}(\hat{\mathbf{X}}_\theta(\mathbf{Z}), \mathbb V[\mathbf X | \mathbf Z]),
\end{equation}
with the estimate of the covariance $\mathbb V[\mathbf X | \mathbf Z]$ using Tweedie's covariance formula \citep{efron2011tweedie}.
After marginalisation of \cref{eq:p_y_z}, this leads to 
\begin{equation}
    p(\mathbf Y | \mathbf Z) \approx \mathcal{N}(\mathcal A, A  \mathbb V[\mathbf X | \mathbf Z]  A^\top +\delta ^2 \mathbf I),
\end{equation}
where $\mathcal A\equiv\mathcal A(\hat{\mathbf X}_\theta(\mathbf Z))$, and $A\equiv\nabla\!_{\hat{\mathbf X}} \mathcal A$, i.e. the Jacobian of the non-linear function $\mathcal A$ w.r.t. $\hat{\mathbf X}_\theta(\mathbf Z)$. 
For more details about MMPS, we refer to \cref{app:mmps}. The exact computation of the likelihood score via Conjugate Gradient is summarized in \cref{alg:mmps}.

\textbf{Diffusion forcing and MMPS are complementary.}
Diffusion forcing enables stable arbitrary-length rollouts by training with independent per-frame noise levels, preventing distribution shift when mixing clean context frames with noisy future frames~\citep{chen2024diffusion}.
MMPS enables zero-shot task generalization by steering the generation toward task-specific observations at inference time.
At high noise levels, when $\mathbb{V}[\mathbf{X}|\mathbf{Z}]$ is large and $\Sigma_k^{-1}$ small, the prior dominates over the small likelihood score. When the noise decreases and predictions are more reliable, the guidance becomes stronger. Together, they allow \ourMethod to act as a single foundation model across all tasks without fine-tuning.

\subsection{Task-Specific Operators}\label{sec:tasks}

\begin{figure*}[t]
    \centering
    \includegraphics[width=\linewidth]{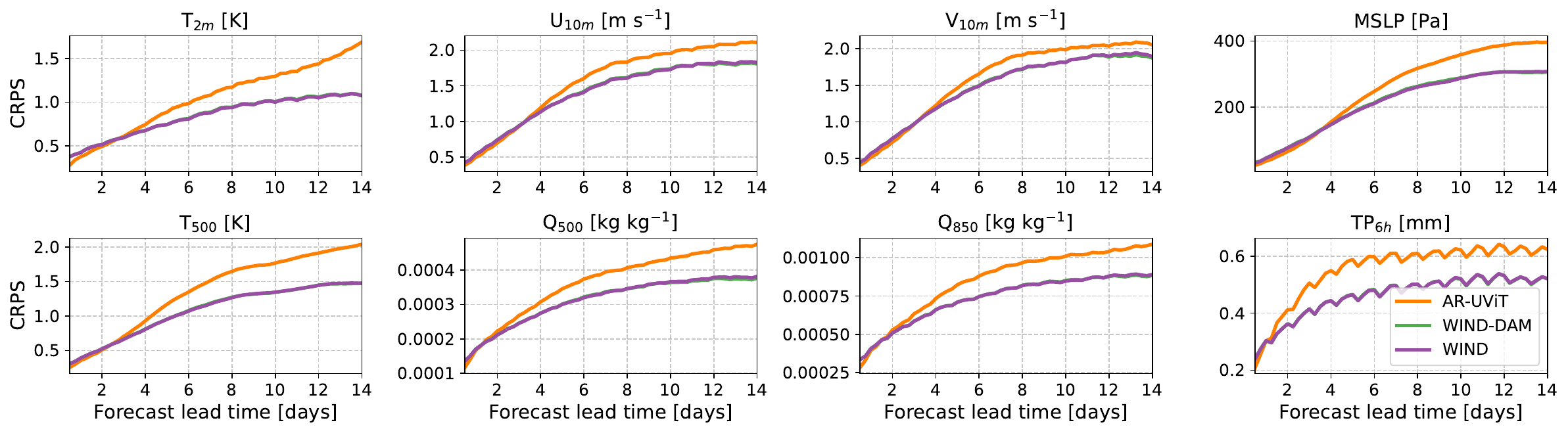}
    \caption{\textbf{Probabilistic forecast performance} 
    We evaluate the 14-day forecast skill using the CRPS (lower is better) averaged over 100 initializations in 2021. We compare the unconstrained \ourMethod baseline against the performance with enforced dry air mass conservation. The large overlap demonstrates that the physics constraint does not degrade the probabilistic forecast skill. \ourMethod outperforms the autoregressive AR-UViT baseline within only a few days.}
    \label{fig:crps}
\end{figure*}

The forward operator $\mathcal{A}$ fully specifies the downstream application. All tasks share the identical pre-trained model and inference loop, only $\mathcal{A}$ changes. The complete inference algorithms for probabilistic forecasting and for the inverse problems are formalized in \cref{alg:wind_forecasting,alg:wind_inverse}.

At each inference step, the model predicts the clean atmospheric state $\hat{\mathbf{X}}_\theta(\mathbf{Z})$, the $\mathcal{A}$ operator is applied to this prediction: $\hat{\mathbf{Y}} = \mathcal{A}(\hat{\mathbf{X}}_\theta(\mathbf{Z}))$, and the likelihood gradient is computed with respect to the target measurement $\mathbf{Y} =\mathcal{A}(\mathbf{X})$. Crucially, because the likelihood gradient is backpropagated through $\mathcal{A}$ into the full atmospheric state $\mathbf{Z}$, the reverse diffusion process always operates at the native resolution of $\mathbf{X}$, regardless of the dimensionality of $\mathbf{Y}$.

\textbf{i) Probabilistic forecasting:} $\mathcal{A} = \mathbf{I}$ (identity, no guidance). One clean frame initializes the forecast and all future frames are denoised by the prior alone.

\textbf{ii) Spatial downscaling}:
$\mathcal{A}(\mathbf{X}) = \mathrm{AvgPool}_{s \times s}(\mathbf{X})$.
The generated high-resolution reconstruction must be consistent with the low-resolution, spatially averaged input.

\textbf{iii) Temporal downscaling:}
$\mathcal{A}(\mathbf{X}) = \frac{1}{T}\sum_{t=1}^T \mathbf{x}^t$.
The daily-mean observation $\mathbf{Y}$ constrains the temporal average, while the prior generates sub-daily variability.

\textbf{iv) Sparse reconstruction:}
$\mathcal{A}(\mathbf{X}) = \mathbf{M} \odot \mathbf{X}$,
where $\mathbf{M}$ is a binary sensor mask. The prior coherently fills unobserved regions independently of sensor geometry.

\textbf{v) Global dry air mass conservation:}
$\mathcal{A}(\mathbf{X}) = f_\mathrm{DAM}(\mathbf{x}^t) = c_\mathrm{DAM}$.
This non-linear operator (see \cref{eq:dam_computation}) enforces global mass conservation across channels.

\textbf{vi) Warmer climate scenarios:}
$\mathcal{A}(\mathbf{X})_c = \frac{1}{HW}\sum_{h,w} x^t_{c,h,w}$.
Constraining the global spatial mean of thermodynamic channels.

\section{Experiments} \label{Sec:experiments}

\paragraph{Probabilistic forecasting}
Probabilistic forecasting is a fundamental benchmark to validate that \ourMethod has learned a robust, physically consistent prior of atmospheric dynamics. We frame forecasting as a conditional generation task: given a context window of past clean frames $\mathbf{X}_{\text{past}}$, we sample the future frames $\mathbf{X}_{\text{future}}$ by initializing them as pure noise and denoising them using our model, as shown on the left side of \cref{fig:figure_2} but without the guidance part. Unlike autoregressive baselines (e.g. GenCast, GraphCast) that optimize specifically for forecasting, our independent noise training formulation allows us to perform forecasting purely at inference time without task-specific fine-tuning.

We evaluate forecast skill on 24 initial conditions from 2021, generating 10-member ensembles for a 14-day lead time. We assess performance using the continuous ranked probability score (CRPS) for accuracy and the spread-skill ratio (SSR) for calibration. Our autoregressive AR-UViT baseline replicates the GenCast forecasting setting. Full details regarding the evaluation metrics and baseline configurations are provided in \cref{app:forecasting_details}.

\begin{figure*}[t]
    \centering
    \includegraphics[width=0.99\linewidth]{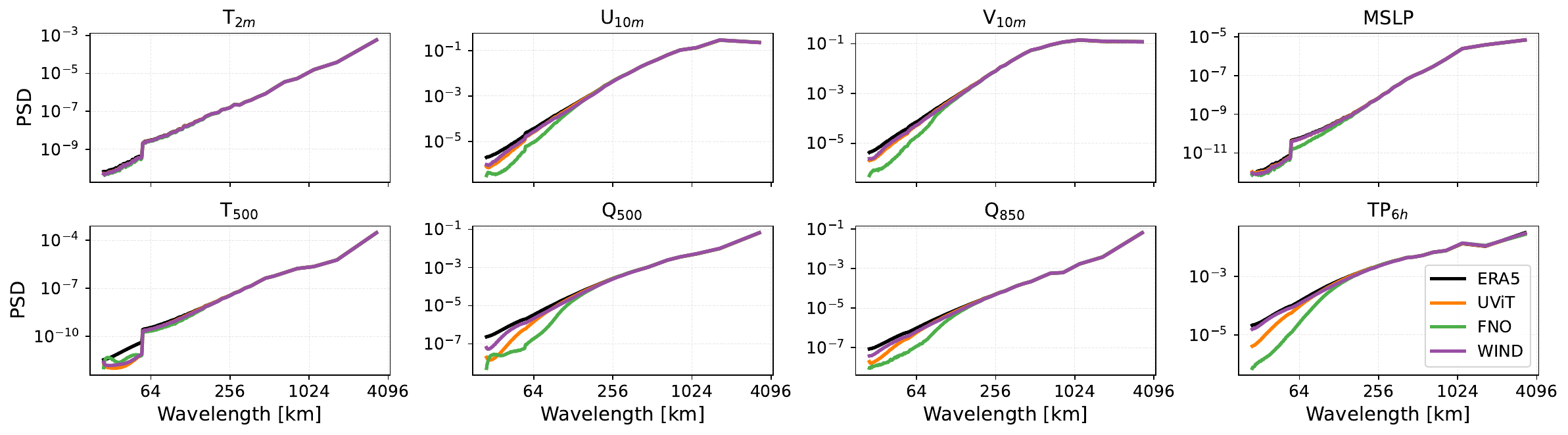}
    \caption{\textbf{Power spectra for spatial downscaling.} We compare the PSD of the ERA5 ground truth, a specialized FNO and UViT model, and \ourMethod. 
    \ourMethod closely tracks the energy spectrum of ERA5 across all scales, preserving high-frequency details. In contrast, the deterministic FNO baseline exhibit spectral drop-off at high frequencies. While UViT performs on par with our method for surface variables, it struggles with the atmospheric variables $Q$ and $\text{TP}$.} 
    \label{fig:spatial_downscaling_psd}
\end{figure*}

\cref{fig:crps} shows the CRPS for key atmospheric variables. \ourMethod consistently outperforms the autoregressive AR-UViT baseline in CRPS after the first few days, demonstrating superior stability over longer horizons. In terms of calibration, we analyze the spread-skill ratio (SSR) in \cref{fig:forecasting_ssr}. Ideally, the SSR should be close to 1.0, indicating that the ensemble spread accurately reflects the forecast error. We observe that \ourMethod approaches an SSR of 1.0 over time, transitioning from an initially over-confident state to smoothly saturate at the climatological variance after two weeks as fundamental predictability limits are reached \citep{bauer2015quiet}. Crucially, our method avoids the instability of the AR-UViT baseline, which is also initially under-confident but then tends to overshoot (SSR $>$1) for moisture-related variables (e.g. $Q_{500}$, $Q_{850}$ and $\text{TP}_{6h}$). 
We also conducted a comprehensive comparison against state-of-the-art forecast models on WeatherBench2 \citep{rasp2024weatherbench}, as shown in \cref{tab:scorecard}. While these specialized benchmarks achieve lower absolute CRPS for medium-range forecasting, we emphasize that WIND is trained at a coarser 1.5° resolution to learn a generalizable physical prior rather than maximizing performance on a single task. 

To evaluate the stability over long rollouts, we conduct a 20-year unconstrained rollout initialized in the year 2000.
We compare the physical consistency of WIND against an autoregressive diffusion-based baseline, AR-UViT5.
Both share the same architecture and sequence length $T$.
AR-UViT5 is trained as full sequence diffusion: it receives a clean initial frame as context and is optimized to denoise subsequent frames with a uniform noise level, see \cref{app:forecasting_details}. 
\cref{fig:20y_forecast_psd_latlon_mean_hist} shows that AR-UViT5 exhibits unphysical spikes across all variables, while WIND maintains physical consistency across the entire spectral range.

\paragraph{Spatial downscaling}
Spatial downscaling (increasing the resolution) is critical for impact modeling, as climate model projections have coarse resolutions due to computational constraints. While standard reanalysis products like ERA5 are available at 0.25°, long-term climate projections (e.g. CMIP6) are limited to much coarser grids, which smooth out extreme events. Bridging this resolution gap is essential for local risk assessment.

We frame this as an inverse problem, aiming to recover a high-resolution sequence $\mathbf{X} \in \mathbb{R}^{T \times C \times H \times W}$ from a coarsened observation $\mathbf{Y} \in \mathbb{R}^{T \times C \times (H/s) \times (W/s)}$ generated by a pooling operator $\mathcal{A}(\mathbf{X}) = \text{AvgPool}_{s \times s}(\mathbf{X})$. 
The right panel of \cref{fig:figure_2} illustrates the inference guidance for a single frame.
We benchmark \ourMethod against two specialized deterministic models, a Fourier Neural Operator (FNO) \citep{li2021fno} and a UViT \citep{hoogeboom2023simple} based model on a $s=4$ downscaling task for the entire year of 2021 (see \cref{app:architecture_hp}). In principle, we can downscale data from any resolution to match the resolution of our training data.

Downscaling must recover the small-scale physics which is absent in the low-resolution input. 
As shown in \cref{fig:spatial_downscaling_psd}, \ourMethod demonstrates superior capability in reconstructing these high-frequency details. The power spectral density (PSD) plots reveal that our model  maintains energy levels consistent with the ERA5 ground truth even at the smallest wavelengths, outperforming our UViT baseline. In contrast, the deterministic FNO baseline suffers from spectral bias, exhibiting a sharp drop-off at high frequencies which results in overly smooth, blurry predictions, a known limitation of regression-based objectives. We provide a quantitative comparison in \cref{tab:spatial_downscaling_rmse_comparison} 
and \cref{fig:spatial_downscaling_rmse_normalized}. The specialized UViT baseline achieves the lowest RMSE. This is expected, as deterministic models optimized for pixel-wise error inherently induce spatial smoothing under uncertainty, artificially inflating skill at the expense of physical realism \citep{subich2025fixing}.

Generally, it is expected that a model with the same capacity can perform better when fine-tuned to a specific task, especially in the case of downscaling, where the model can learn to only model the high-frequency patterns during training and rely on the low-resolution condition. However, despite not being trained for this task, \ourMethod remains competitive with (and often outperforms) the deterministic FNO baseline in terms of RMSE. While the specialized UViT baseline achieves slightly lower pixel-wise RMSE, our model outperforms it in capturing high-frequency variability for key atmospheric variables like for example specific humidity.

\paragraph{Temporal downscaling} 
Although Earth system models run at fine temporal scales, outputs are frequently aggregated and archived at daily or monthly resolutions to manage data volumes. This temporal aggregation can obscures the critical dynamics of extreme events. Despite its importance, temporal downscaling remains largely unexplored compared to its spatial counterpart, with only recent works leveraging generative models to bridge this gap \citep{bassetti2024diffesm, schmidt2025generative}. We address this by recovering high-frequency sequences  $\mathbf{X} \in \mathbb{R}^{T \times C \times H \times W}$ from temporally aggregated observations $\mathbf{Y} \in \mathbb{R}^{1 \times C \times H \times W}$  (e.g. daily means). We solve temporal downscaling as a pure inference problem via MMPS guidance, ensuring the generated sequence remains consistent with the daily average while generating plausible sub-daily dynamics (see \cref{app:temporal_details}).

We compare our approach against a specialized model trained specifically for temporal downscaling. We use a specialized UViT trained to downscale the daily average field $\mathbf{Y} \in \mathbb{R}^{1 \times C \times H \times W}$ into four 6-hourly frames  $\mathbf{X} \in \mathbb{R}^{4 \times C \times H \times W}$.

The primary goal of temporal downscaling is to generate plausible sub-daily dynamics that are absent in the coarser temporally average resolution. As shown in \cref{fig:temporal_downscaling_psd_latlon_mean_hist}, \ourMethod excels at this task. The power spectral density plots show that our model recovers the full energy spectrum of the 6-hourly data, matching the ERA5 ground truth across all wavelengths. This is important for variables like precipitation and surface wind, where the daily mean largely smooths out the variance. For total precipitation our model outperforms the specialized UViT baseline. 
The histograms further confirm that \ourMethod and the UViT baseline correctly reproduce the sub-daily distributions. This includes the heavy tails of extreme events, which are essential for risk assessment but typically lost in temporally aggregated data.

\paragraph{Sparse reconstruction}
Global atmospheric reanalysis datasets combine simulations with sparse observations from satellites, weather balloons, and ground stations using data assimilation techniques \citep{hersbach2020era5}. We address the challenge of reconstructing full global fields from spatially disjoint measurements, a critical task for historical reanalysis and filling gaps in satellite data.

We frame this again as an inverse problem: recovering the full state $\mathbf{X}$ from sparse observations $\mathbf{Y}$ defined by a binary masking operator $\mathcal{A}(\mathbf{X}) = \mathbf{M} \odot \mathbf{X}$. Unlike statistical interpolation methods (e.g., Kriging) or specialized models, MMPS allows us to handle arbitrary sensor configurations without retraining (see \cref{app:sparse_details}).
While traditional statistical methods like Kriging minimize error by regressing to the mean, resulting in overly smooth fields, \ourMethod preserves the high-frequency power spectrum of the atmosphere, generating realistic textures even in unobserved regions. Due to the prohibitive $O(N^3)$ computational scaling of Kriging with respect to the number of observation points, we limit its evaluation to a representative day rather than the full evaluation year. As a minimum-variance estimator, Kriging is theoretically guaranteed to produce smoothed fields, making it inherently unsuitable for recovering high-frequency dynamics.

As shown in \cref{tab:rmse_comparison_sparse_1}, as well as \cref{fig:sparse_reconstruction_rmse_1_percent} and \cref{fig:sparse_reconstruction_rmse_10_percent}, \ourMethod outperforms the specialized UViT baseline at reconstruction accuracy for the majority of atmospheric variables. Our model excels at predicting large-scale dynamical fields, delivering a substantial reduction in RMSE for metrics such as geopotential and MSLP. This is in contrast to the downscaling tasks, where the specialized model had lower overall RMSE. Here, the specialized UViT struggles to generalize from the extremely sparse 1\% input. In contrast, \ourMethod leverages its prior of the atmosphere to fill in 99\% unobserved regions coherently, demonstrating that a strong foundation model can outperform specialized training in data-scarce regimes.

The visual comparison in \cref{fig:sparse_reconstruction_example_0_1} (or in \cref{fig:sparse_reconstruction_example_0_01} with 1\% sparsity) highlights the structural advantages of generative models for the task. Kriging indeed produces an overly smooth field that miss high-frequency weather dynamics.  Visually \ourMethod and UViT both generate sharp, realistic looking fields that are indistinguishable from the ERA5 ground truth. The spectral analysis (\cref{fig:sparse_reconstruction_psd_latlon_mean_hist_1_percent}), confirms that \ourMethod closely tracks the energy spectrum of the ground truth across all scales, whereas UViT suffers from smoothing at high frequencies for some fields like precipitation or specific humidity. Both methods closely follow the ERA5 ground truth distribution (\cref{fig:sparse_reconstruction_psd_latlon_mean_hist_1_percent} 
and \cref{fig:sparse_reconstruction_psd_latlon_mean_hist_10_percent}) for most of the variables. However, \ourMethod outperforms the baseline for total precipitation, better covering the extremes events.

\paragraph{Enforcing global dry air mass conservation}

Purely data-driven AI forecasting models tend to become unstable or drift into unphysical states during long rollouts \citep{chattopadhyay2023challenges}. While hybrid architectures like NeuralGCM \citep{kochkov2024neural} try to mitigate this long-term instability by coupling a physics-based dynamical core with neural network parameterizations, other recent AI-based approaches rely on external corrective schemes to explicitly enforce the conservation of global energy, moisture budget and dry air mass \citep{sha2025improving}. We demonstrate that \ourMethod can maintain physical consistency and enforce the global conservation of dry air mass purely at inference time, treating it strictly as an inverse problem (see \cref{app:conservation_details}). We define a global operator $A_{\text{DAM}}$ that computes the global integral of dry air mass (DAM). Using MMPS guidance, we constrain the generation process to satisfy $A_{\text{DAM}}(X) = C_{\text{DAM}}$ at every step (see \cref{eq:dam_computation}).

\begin{figure}[t!] 
\centering 
\includegraphics[width=0.99\linewidth]{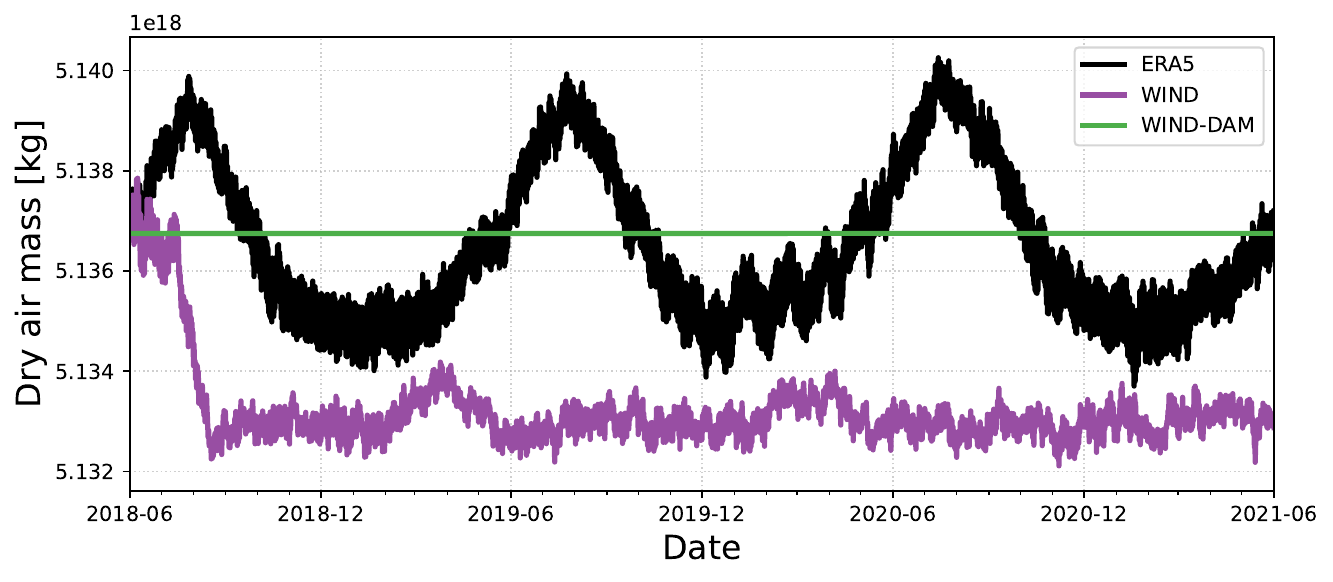} \caption{\textbf{Stability of global dry air mass conservation.} The DAM of \ourMethod without constraint drifts around 200 forecast days. ERA5 ground truth shows a seasonal cycle. \ourMethod with DAM guidance strictly enforces conservation for the entire 4-year rollout.} 
\label{fig:dam_time}
\end{figure}

We evaluate the stability of our method on a 4-year rollout. As shown in \cref{fig:dam_time}, a standard free run eventually drifts after 200 days, after initially mimicking the unconstrained fluctuations present in the ERA5 training data. On the other hand, the MMPS-guided run strictly maintains the global DAM at the target value for the entire duration, demonstrating that our framework can correct unphysical drifts for long rollouts without retraining. Importantly, \cref{fig:crps} confirms that the global DAM conservation does not degrade short-term forecast skill across 100 initializations in 2021. While precipitation skill does not improve, this aligns with \citet{sha2025improving}, who attributes such improvements to moisture and energy constraints. Enforcing these constraints requires surface flux variables that are missing from our dataset.

\paragraph{Applying the model in a different climate}

Evaluating global, data-driven AI models under out-of-distribution (OOD) climate scenarios is a major challenge. When initialized in a warmer state, unconstrained AI models inherently drift back to their training climatology during autoregressive rollouts, dissipating the prescribed thermodynamic anomalies \citep{rackow2024robustness}. To overcome this unphysical drift, we propose a generative surrogate scenario to stress-test the model's learned prior. We perturb the initial conditions and use the $\mathcal{A}$ operator to continuously constrain only the global mean thermodynamic state. Crucially, by constraining only the global mean rather than individual grid cells, the model retains the spatial freedom to naturally simulate localized weather dynamics.

We apply this to Storm Bernd (July 2021, Germany), assuming a idealized +2K global warming and +14\% specific humidity increase via Clausius-Clapeyron scaling \citep{trenberth2003changing}. While explicitly prescribing this constant offset alters the global thermodynamic response, it acts as a necessary form of thermodynamic conditioning, to sustain the forcing within the AI model. Although this uniform perturbation is a simplified alternative to spatially varying climate change deltas derived from Earth system models \citep{duan2025testing}, it effectively establishes a warmer background climate and prevents the model from reverting to its training distribution. This allows the model's learned prior to internally resolve localized extreme weather responses under OOD conditions.

To isolate the impact of our guidance mechanism, we conduct an ablation study with three scenarios 
(see \cref{app:storyline_details}). We run 24-hour forecasts (4 frames) comparing (i) a \textbf{control} run starting from the unperturbed initial condition, (ii) a \textbf{warm free} run, where only the initial state is perturbed (warmer and wetter) but the model evolves unconstrained and (iii) a \textbf{warm guided} run, with the perturbed initial conditions and where MMPS actively enforces the mean thermodynamic constraint at every step.

\begin{figure*}
    \centering
    \includegraphics[width=\linewidth]{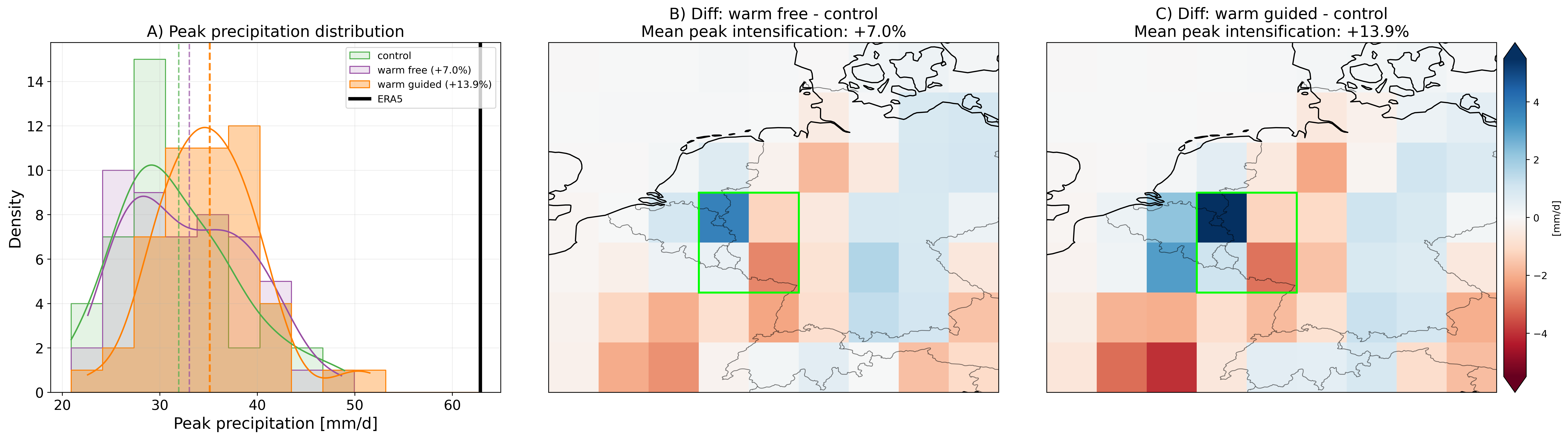}
    \caption{\textbf{Ablation of thermodynamic guidance for storm Bernd.} 
    \textbf{(A)} Peak 24-hour precipitation density ($N=50$) within the target region (green box). Dashed lines denote ensemble means; solid black is ERA5. 
    \textbf{(B-C)} Spatial difference maps relative to control. The unconstrained warm free run \textbf{(B)} dissipates the thermodynamic signal, whereas active guidance \textbf{(C)} sustains robust intensification.}
    \label{fig:storyline_results}
\end{figure*}

As shown in \cref{fig:storyline_results}, the warm guided ensemble effectively sustains the out-of-distribution thermodynamic shift. To compute the intensification, we average the pairwise differences of the peak precipitation pixels between the control and the warm guided runs over the target region ($49^{\circ}$-$52^{\circ}$N, $5.5^{\circ}$-$8.5^{\circ}$E). Across the 50 ensemble members, we observe a mean peak intensification of +13.9\%, closely matching the theoretical Clausius-Clapeyron baseline expected for a +2K warming ($\sim$14\%).

Crucially, the unconstrained warm free ablation run retained only 50.3\% of this signal (+7.0\%). Without active guidance, the model's learned prior effectively diffuses the initial state anomaly back toward its training climatology \citep{rackow2024robustness}, leading to a significant underestimation of the extreme event. Meanwhile, our guided framework maintains excellent structural stationarity: for the 500 hPa geopotential height, we observe a structural similarity (SSIM) of $>0.98$ and zero-pixel displacement of the storm center, while horizontal wind speeds changed by less than 1\%.

Finally, while the ensemble mean aligns with theoretical expectations, the spatial response remains highly heterogeneous, which is consistent with observational findings \citep{traxl2021role}. Our generative ensemble approach reveals the importance of capturing these localized tail risks. In the worst-case guided realization (Member 48), we observed a local precipitation increase of +56.9\% (+18.7 mm/d), whereas the warm free run produced a maximum increase of only +2 mm/d. This behavior aligns with literature finding that localized extreme precipitation often exhibits Super-Clausius-Clapeyron scaling, exceeding the $\sim$14\% baseline due to complex non-linear feedbacks \citep{berg2013strong}.

\section{Discussion and Future Work}

In this work, we presented \ourMethod, a framework that marks a shift from specialized, single-task models to a unified probabilistic foundation model of the atmosphere. Our results highlight an inherent trade-off between task-specific specialization and flexibility. While specialized baselines of comparable size achieve slightly lower RMSE in our experiments, we found that our method often exhibits less spectral smoothing and preserves more high-frequency details (\cref{fig:spatial_downscaling_psd,fig:temporal_downscaling_psd_latlon_mean_hist}). By decoupling atmospheric dynamics from task specific objectives, \ourMethod supports arbitrary inference constraints, such as enforcing physical consistency or reconstructing non-uniform sensor inputs. We even outperform our specialized baselines for sparse reconstruction and forecasting. We argue that this versatility outweighs the marginal pixel-wise gains of single-task models. Diffusion forcing enables \ourMethod to substantially increase autoregressive stability compared to full sequence diffusion (\cref{fig:20y_forecast_psd_latlon_mean_hist}). Generating two decades of data requires over 7,300 autoregressive steps, demonstrates the robustness of our prior.

Our experiments demonstrate that active guidance is essential for out-of-distribution evaluation, as the unconstrained model quickly regresses toward the training distribution \citep{rackow2024robustness}. To counteract this drift, recent AI adaptations have relied on explicit dynamic nudging or costly ensemble filtering \citep{duan2025testing}, whereas our framework sustains a prescribed thermodynamic background state while leaving local dynamics unconstrained. We emphasize that this single-event analysis serves as a proof of concept. Localized extremes should be interpreted as plausible tail risks of the learned distribution rather than strict causal predictions.

Because AI weather models generally lack physical constraints, they are prone to violating fundamental conservation laws, which in turn drives unphysical drift and error accumulation during autoregressive rollouts \citep{sha2025improving}. \ourMethod successfully enforces global integral constraints (e.g. dry air mass) purely at inference time. In the inverse problem formulation, the operator $\mathcal{A}$ is not restricted to linear functions, any differentiable  operator can be introduced at inference time without retraining, as demonstrated by the non-linear dry air mass conservation operator (\cref{eq:dam_computation}). While\ourMethod stabilizes global budgets, it does currently not enforce local conservation laws. Achieving strict local conservation remains a challenge for AI models, representing an important direction for future research.

WIND learns relationships for downscaling exclusively from ERA5, however it can also be applied to climate model projections. While biases in climate models introduce domain shifts, these can be mitigated using established two-stage bias correction methods \citep{aich2026conditional, wan2023debias}. Unlike state-of-the-art frame-wise downscaling methods \citep{aich2026conditional, hess2025fast}, \ourMethod also captures temporal correlations in high-frequency patterns. By training with larger window sizes, \ourMethod can be used to downscale monthly data to daily resolution.

Despite promising results, our approach has limitations that we leave for future research. The primary drawback is inference speed. The iterative denoising and MMPS gradient calculations result in a high computational cost (\cref{tab:runtime}). This can be addressed via distillation techniques \citep{salimans2022progressive, sabour2025align, potaptchik2026meta}. Additionally, while the model captures the relative intensification of events, it underestimates extreme events like Storm Bernd. This is likely a combination of our coarse 1.5° spatial resolution and the general difficulty of out-of-distribution generalization of data-driven models \citep{sun2025can}. By training at 0.25° resolution, \ourMethod could better resolve these extremes.

Future iterations could incorporates sea surface temperatures and energy fluxes to capture longer-range phenomena (e.g. the El Niño Southern Oscillation) and enforce thermodynamic budgets. The additional information could improve the model's forecast performance \citep{sha2025improving}. Moreover, our inference framework offers to move beyond reanalysis data by directly assimilating observations from satellites and stations \citep{andry2025appa, yang2025generative}. 

By learning a robust probabilistic prior of atmospheric dynamics, \ourMethod provides a unified framework that flexibly adapts to diverse downstream tasks, marking a step towards a general-purpose atmospheric AI model.

\section*{Acknowledgements}

The ELLIS Unit Linz, the LIT AI Lab, the Institute for Machine Learning, are supported by the Federal State Upper Austria. We thank the projects FWF AIRI FG 9-N (10.55776/FG9), AI4GreenHeatingGrids (FFG- 899943), Stars4Waters (HORIZON-CL6-2021-CLIMATE-01-01), FWF Bilateral Artificial Intelligence (10.55776/COE12). We thank NXAI GmbH, Audi AG, Silicon Austria Labs (SAL), Merck Healthcare KGaA, GLS (Univ. Waterloo), T\"{U}V Holding GmbH, Software Competence Center Hagenberg GmbH, dSPACE GmbH, TRUMPF SE + Co. KG. Funded under the Michael Aich acknowledges funding from the Excellence Strategy of the Federal Government and the L\"ander through the TUM Innovation Network EarthCare. This is a Past to Future contribution; The Past to Future (P2F) project has received funding from the European Union’s Horizon Europe research and innovation programme under grant agreement No. 101184070: Funded by the European Union. 
Niklas Boers acknowledges funding by the Volkswagen Foundation. This is ClimTip contribution \#167; the ClimTip project has received funding from the European Union's Horizon Europe research and innovation programme under grant agreement No. 101137601: Funded by the European Union. Views and opinions expressed are however those of the author(s) only and do not necessarily reflect those of the European Union or the European Climate, Infrastructure and Environment Executive Agency (CINEA). Neither the European Union nor the granting authority can be held responsible for them.

\section*{Impact Statement}

This work aids climate adaptation and renewable energy planning by introducing a unified, computationally efficient foundation model for analyzing atmospheric dynamics and extreme weather events. By enabling ensemble based weather generation and probabilistic forecasting, our framework empowers policymakers and scientists to better explore the increasing risks associated with global warming. However, reliance on generative models for safety for impact assessments requires rigorous validation to mitigate the risks of hallucinated physics or biases.

\bibliography{refs}
\bibliographystyle{icml2026}

\newpage
\appendix
\counterwithin{figure}{section}
\counterwithin{table}{section}

\clearpage

\section{Experimental Details}

\subsection{Dataset}\label{Appendix_dataset}

\paragraph{ERA5} We use the ERA5-dataset \citep{hersbach2020era5}, a state-of-the-art reanalysis dataset provided by the European Center for Medium-Range Weather Forecasting (ECMWF). We use the 1.5° resolution of the data (with a grid size of 240 $\times $121 pixel) at 6-hourly temporal resolution provided by Weatherbench-2 \citep{rasp2024weatherbench}. We include 70 prognostic variables (input and output): \textbf{Surface Variables} (5): Precipitation (p), 2m temperature (2t), mean sea level pressure (mslp), and the u and v components of 10m wind (10u, 10v). \textbf{Pressure Level Variables} (65): Temperature (t), geopotential (z), specific humidity (q), and the u and v components of wind (u, v) across 13 distinct pressure levels.

\paragraph{Pre-processing}
We normalize each variable, at each pressure level to zero mean and unit variance using statistics calculated over the training set. For precipitation variables, which exhibit heavy-tailed distributions with values spanning multiple orders of magnitude, we apply a log-transformation prior to the standard normalization following \citet{aich2025diffusion,hess2025fast}. Specifically, we map the raw precipitation values $x$ (in m) to $x=\log_{10}(1000x+1)$. The factor 1000 converts the values to millimeters, and the constant 1 ensures numerical stability for zero-precipitation regions. We further incorporate auxiliary static and dynamic features into the input state. Static features include the land-sea mask, soil type, and surface geopotential. The surface geopotential is normalized to zero mean and unit variance, while the masks are kept as binary identifiers. To preserve the spherical geometry of the Earth without discontinuities at the date line, we embed the spatial grid coordinates $(\phi,\lambda)$ into 3D Cartesian space as $(\sin\phi,\cos\phi,\cos\phi \sin\lambda)$ . Furthermore, to explicitly encode the temporal cycle, we append sine and cosine embeddings of both the annual cycle (year progress) and the diurnal cycle (local time of day, derived from UTC time and longitude) to the input channels

\subsection{Architecture \& Hyperparameters} \label{app:architecture_hp}

\paragraph{Architecture} 
For our model and all UViT baselines we use the same UViT \citep{hoogeboom2025simpler} backbone and use adaptions of the Transformer layers as done in \citet{song2025history} to adapt the model to temporal data.
For the downstream tasks we modified the training paradigm (deterministic / diffusion) and training setup (input and outputs) when necessary. 

We configure the UViT with a hierarchical structure across four spatial resolutions. Starting from a base channel dimension of 256, the channels progress as (256, 512, 1024, 2048) across resolutions. After initial patchification with patch size 2, we employ a hybrid block design to balance computational efficiency with global context modeling. Specifically, we use four residual blocks \citep{he2016resnet} at each of the two highest spatial resolutions, and four Transformer blocks \citep{vaswani2017attention} at each of the two lowest resolutions. Each Transformer block employs 4 attention heads and 1D rotary positional embeddings (RoPE) \citep{su2021roformer} applied across both spatial and temporal dimensions. This configuration results in a total parameter count of approximately 458 million.
We utilize a temporal window of $T=5$ in order to predict a full day at 6-hourly resolution given one state. Each timestep comprisies 70 atmospheric variables at a resolution of 240 $\times$ 121 pixels. 

For the downscaling we additionally use a FNO \citep{li2021fno} as an additional baseline. 
We use $L=8$ layers, a hidden width of $128$, and a grid positional embedding, and we truncate spectral convolutions at $k_{max} = [64, 128]$ modes. This results in a total parameter count of approximately 545 million.

\paragraph{Hyperparameters} 
We use a cosine learning rate schedule with a linear warmup period of 5 epochs, reaching a peak learning rate of $1\times10^{-4}$ before decaying to a minimum of $1\times10^{-6}$. The model is trained for a maximum of 60 epochs with a global effective batch size of 16. We use the Adam, clipping gradients above a norm of 0.8. To improve training efficiency, we utilize bfloat16 mixed precision. During evaluation, we use an EMA version of the model, with an EMA decay rate of 0.999. Since the native ERA5 spatial grid of 240$\times$121 (longitude $\times$ latitude) does not result in an even number after multiple downsampling stages, we bilinearly interpolate the input to a 240$\times$128 grid before the network and interpolate it back to the original resolution after the final layer.

During inference, all our diffusion based models use DDIM sampling with a 15 deterministic steps (thus $\eta = 0$). 
When using MMPS for guidance, we use 2 iteration steps for the conjugate gradient method and a noise variance $\delta^2 = 0.0015$.

\begin{algorithm*}[b]
\caption{MMPS likelihood score}
\label{alg:mmps}
\begin{algorithmic}
    \STATE {\bfseries Input:} $\hat{\mathbf{X}}_\theta(\mathbf{Z})$, observation $\mathbf{Y}$, operator $\mathcal{A}$, noise level $k$, task noise $\delta^2$, CG iterations $M$
    \STATE $\hat{\mathbf{Y}} \gets \mathcal{A}(\hat{\mathbf{X}}_\theta(\mathbf{Z}))$ \COMMENT{Apply task operator to get prediction of $\mathbf{Y}$ given the current estimate $\hat{\mathbf{X}}_\theta(\mathbf{Z})$}
    \STATE $A \gets \nabla\!_{\hat{\mathbf{X}}} \mathcal{A}(\hat{\mathbf{X}}_\theta(\mathbf{Z}))$ \COMMENT{Calculate Jacobian of $\mathcal{A}$ at current estimate $\hat{\mathbf{X}}_\theta(\mathbf{Z})$}
    \STATE $\Sigma_k \gets \beta(k)^2\, A\,(\nabla\!_{\mathbf{Z}}\hat{\mathbf{X}}_\theta(\mathbf{Z}))^\top A^\top + \delta^2\mathbf{I}$ \COMMENT{Tweedie covariance; $\nabla\!_{\mathbf{Z}}\hat{\mathbf{X}}_\theta(\mathbf{Z})$ never explicitly evaluated}
    \STATE $\mathbf{v} \gets \mathrm{CG}(\Sigma_k,\; \mathbf{Y} - \hat{\mathbf{Y}},\; M)$ \COMMENT{Solve $\Sigma_k \mathbf{v} = \mathbf{Y} - \hat{\mathbf{Y}}$ with CG; computed via vector-Jacobian product}
    \STATE \textbf{return} $(\nabla\!_{\mathbf{Z}}\hat{\mathbf{X}}_\theta(\mathbf{Z}))^\top A^\top \mathbf{v}$ \COMMENT{$\approx \nabla\!_{\mathbf{Z}} \log p(\mathbf{Y}|\mathbf{Z})$; computed via vector-Jacobian product}
\end{algorithmic}
\end{algorithm*}

\section{Approach Details} \label{app:dm_approach_details}

\subsection{Diffusion Models Background} \label{app:diffusion_background}

Diffusion models define a stochastic \emph{forward} process that corrupts clean data $\mathbf{x} \in \mathbb{R}^D$ by gradually adding noise over a continuous time interval $k \in [0,1]$. This process can be modeled as a Stochastic Differential Equation (SDE) \citep{song2020score}:
\begin{align}
    d\mathbf{z}_k = \mathbf{f}(\mathbf{z}_k, k)dk + g(k) d\mathbf{w}
\end{align}
where $\mathbf{w}$ is a standard Wiener process. The drift $\mathbf{f}(\mathbf{z}_k, k)$ and diffusion coefficient $g(k)$ are chosen such that $\mathbf{z}_0 = \mathbf{x}$ (clean data) and $\mathbf{z}_1$ approaches a standard Gaussian distribution.

To generate samples, we simulate the \emph{reverse} process \citep{anderson1982reverse} which runs backward in time from $k=1$ to $k=0$:
\begin{align}
    d\mathbf{z}_k = [\mathbf{f}(\mathbf{z}_k, k) - g(k)^2 \nabla_{\mathbf{z}_k} \log p_k(\mathbf{z}_k)]dk + g(k) d\tilde{\mathbf{w}}
\end{align}
where $d\tilde{\mathbf{w}}$ is a reverse-time Wiener process. Since the true score function $\nabla_{\mathbf{z}_k} \log p_k(\mathbf{z}_k)$ is intractable, it is approximated by a neural network $\mathbf{s}_\theta(\mathbf{z}_k, k) \approx \nabla_{\mathbf{z}_k} \log p_k(\mathbf{z}_k)$, trained via denoising score matching \citep{song2020score, vincent2011connection}.

\subsection{Diffusion Forcing Training Details} \label{app:diffuion_forcing_details}

\begin{algorithm*}[t]
\caption{WIND forecasting}
\label{alg:wind_forecasting}
\begin{algorithmic}
    \STATE {\bfseries Input:} Clean context frame $\mathbf{x}^0$, sequence length $T$, DDIM steps $N$, noise schedule $\{k_n\}$, stochasticity $\eta$
    \STATE {\bfseries Additional input (for guided only):} Observation $\mathbf{Y}$, operator $\mathcal{A}$, task noise $\delta^2$, CG iterations $M$
    \FOR{$t = 1, \dots, T$}
        \STATE $\mathbf{z}^t \sim \mathcal{N}(\mathbf{0}, \mathbf{I})$ \COMMENT{Initialise future frames as pure noise}
    \ENDFOR
    \STATE $\mathbf{Z} \gets \{\mathbf{z}^0 {=} \mathbf{x}^0,\, \mathbf{z}^1, \dots, \mathbf{z}^T\}$ \COMMENT{Context frame is clean ($k=0$), future frames are noisy}
    \FOR{$n = N, N{-}1, \dots, 1$}
        \STATE $k \gets k_n$,\quad $k' \gets k_{n-1}$
        \STATE $\hat{\mathbf{X}} \gets \hat{\mathbf{X}}_\theta(\mathbf{Z})$ \COMMENT{Predict clean sequence given clean context frame and noised sequence $\mathbf{Z}$}
        \IF{guided}
            \STATE $\mathbf{s}_\mathrm{lhd} \gets \mathrm{MMPS}(\hat{\mathbf{X}}_\theta(\mathbf{Z}),\, \mathbf{Y},\, \mathcal{A},\, k,\, \delta^2,\, M)$ \COMMENT{Likelihood score via \cref{alg:mmps}}
        \ENDIF
        \FOR{$t = 1, \dots, T$}
            \STATE $\tau \gets 1 - {\alpha(k)^2\,\beta(k')^2} / {\alpha(k')^2\,\beta(k)^2}$
            \STATE $\boldsymbol{\epsilon} \sim \mathcal{N}(\mathbf{0}, \mathbf{I})$
            \STATE $\mathbf s ^t \gets (\mathbf z^t -\alpha(k)\hat{\mathbf x} ^t)/\beta(k)^2$
            \IF{guided}
                \STATE $\mathbf{s}^t \gets \mathbf{s}^t + \mathbf{s}_\mathrm{lhd}^t$ \COMMENT{Posterior guidance correction towards observation $\mathbf{Y}$}
            \ENDIF
            \STATE $\mathbf{z}^t \gets \alpha(k')\,\hat{\mathbf{x}}^t + \beta(k')\beta(k)\sqrt{1{-}\eta\tau}\,\mathbf s^t + \beta(k')\sqrt{\eta\tau}\,\boldsymbol{\epsilon}$ \COMMENT{DDIM update; $\eta{=}0$ deterministic, $\eta{=}1$ stochastic}
        \ENDFOR
    \ENDFOR
    \STATE \textbf{return} $\hat{\mathbf{X}}$ \COMMENT{Unguided: samples from prior $p_{\theta}(\mathbf{X} \mid \mathbf{x}^0)$; guided: approximate posterior samples $p_{\theta}(\mathbf{X} \mid \mathbf{x}^0, \mathbf{Y})$}
\end{algorithmic}
\end{algorithm*}

Standard video diffusion adds noise to all frames at the same rate \citep{ho2022video}. In contrast, following the diffusion forcing paradigm \citep{chen2024diffusion}, we sample a \emph{noise level} $k \in [0,1]$ independently for each frame $t \in \{1, \dots, T\}$. This allows the model to learn to predict any frame given any arbitrary combination of clean or noisy context frames.
For a given frame $\mathbf{x}^t$, let $k^t$ be its sampled noise level. The forward diffusion process is given by:
\begin{equation}
    \mathbf{z}^t = \alpha(k^t)\mathbf{x}^t + \beta(k^t)\boldsymbol{\epsilon}^t, \quad \boldsymbol{\epsilon}^t \sim \mathcal{N}(\mathbf{0}, \mathbf{I}),
\end{equation}
where $\alpha(k^t)$ and $\beta(k^t)$ are the signal and noise schedule coefficients defined by the diffusion time $k^t$. The noised sequence is denoted as $\mathbf{Z} = \{\mathbf{z}^1, \dots, \mathbf{z}^T\}$.

\paragraph{Objective}
The neural network $\hat{\mathbf{X}}_\theta(\mathbf{Z})$ is trained to predict the clean atmospheric state $\mathbf{X} = \{\mathbf{x}^1, \dots, \mathbf{x}^T\}$ given the noised sequence $\mathbf{Z}$. A critical distinction in our approach, motivated by findings from \citet{sun2025noise}, is that we \textbf{do not} condition the network on the noise levels defined by the set of indices $\mathbf{k} = \{k^1, \dots, k^T\}$. The model is strictly a mapping $\hat{\mathbf{X}}_\theta: \mathbb{R}^{T \times C \times H \times W} \to \mathbb{R}^{T \times C \times H \times W}$.
The objective function is the weighted mean squared error,
\begin{equation}
    \mathbb{E}_{\mathbf{X}, \boldsymbol{\epsilon}, \mathbf{k}} \left[ \sum_{t=1}^T \sum_{c=1}^C w_c \sum_{h,w=1}^{H,W} a_{h,w} \| \mathbf{x}^t_{c,h,w} - \hat{\mathbf{X}}_{\theta}(\mathbf{Z})^t_{c,h,w} \|^2_2 \right],
\end{equation}
where $t$ indexes the frame index, $c$ the physical variables (channels) and $(h, w)$ the spatial coordinates. To account for varying grid cell sizes, $a_{h,w}$ represents the normalized cell area such that $\frac{1}{HW} \sum_{h,w} a_{h,w} = 1$. The channel-specific weights $w_c$ are adopted from previous work in atmospheric modeling \citep{lam2022graphcast, price2023gencast}.
We can infer the score function $\mathbf{s}_\theta(\mathbf{Z})$ from the data prediction model $\hat{\mathbf{X}}_\theta(\mathbf{Z})$ \citep{kingma2023elbo} using:
\begin{equation}
    \bfs_\theta(\bfZ)^t = -\beta(k^t)^{-2} \left( \mathbf{z}^t - \alpha(k^t) \hat{\mathbf{X}}_{\theta}(\mathbf{Z})^t \right), 
\end{equation}
where $\bfs_\theta(\bfZ) = \{   \bfs_\theta(\bfZ)^1, \ldots, \bfs_\theta(\bfZ)^T \}$.

We use a rectified noise schedule \citep{liu2023flow,lipman2023flow}, that means our coefficients $\alpha(k^t)=k^t \alpha_{\textrm{min}} + (1-k^t)$ and $\beta(k^t)=k^t + (1-k^t)\beta_{\textrm{min}}$, with $ \alpha_{\textrm{min}}= \beta_{\textrm{min}}=0.001$. We sample each $k^t \sim \mathcal{U}(0, 1)$ independently.

\subsection{Sampling Details} \label{app:sampling_details}

For our forecasting tasks, we provide the model with one clean state and denoise all the others with the same noise level $k$. For all other tasks, we initialize all states noisy and denoise, and just use guidance to reach the desired output.

For frame denoising, we employ DDIM sampling.
Specifically, to transition from noise level $k$ to $k'$ ($k' < k$), we denoise each frame within the window via:
\begin{align}
    \mathbf{z}^t \leftarrow \ & \alpha(k') \hat{\mathbf{X}}_{\theta}(\mathbf{Z})^t + \beta(k') \sqrt{1-\eta \tau} \frac{\mathbf{z}^t - \alpha(k) \hat{\mathbf{X}}_{\theta}(\mathbf{Z})^t}{\beta(k)} \notag \\ 
    + \ & \beta(k') \sqrt{\eta \tau} \mathbf{\epsilon} ,
\end{align}
with $\mathbf{\epsilon} \sim \mathcal{N}(\mathbf{0}, \mathbf{I})$ and
\begin{equation*}
    \tau = 1 - \frac{\alpha(k)^2 \beta(k')^2}{\alpha(k')^2 \beta(k)^2} .
\end{equation*}

Under guidance, we additionally update each $\mathbf{z}^t$ using the likelihood score $\nabla_{\mathbf{Z}} \log p(\mathbf{Y} | \mathbf{Z})$. For further details about how to approximate this likelihood using MMPS, we refer to \cref{app:mmps}.

\subsection{Approximating the Likelihood Using MMPS} \label{app:mmps}

This section is a recap of the MMPS algorithm, more detailed information can be found in \citet{rozet2024learning}.
Moment matching posterior sampling (MMPS) considers a Gaussian approximation:
\begin{equation}
    p(\mathbf{X}|\mathbf{Z}) \approx \mathcal{N}(\hat{\mathbf{X}}_\theta(\mathbf{Z}), \mathbb V[\mathbf X | \mathbf Z]),
\end{equation}
with the estimate of the covariance $\mathbb V[\mathbf X | \mathbf Z]$ using Tweedie's covariance formula \citep{efron2011tweedie}:
\begin{equation}
    \mathbb{V}[\mathbf{X}|\mathbf{Z}] \approx \beta(k)^2\,\nabla\!_{\mathbf{Z}}\hat{\mathbf{X}}_\theta(\mathbf{Z}) ,
\end{equation}
where $\beta(k)^2$ is the noise covariance of the forward diffusion process at level $k$, and $\nabla\!_{\mathbf{Z}}\hat{\mathbf{X}}_\theta(\mathbf{Z}) \equiv\nabla\!_{\mathbf{Z}}\hat{\mathbf{X}}$ is the Jacobian of the denoiser.
Marginalisation of \cref{eq:p_y_z} leads to
\begin{equation}
    p(\mathbf Y | \mathbf Z) \approx \mathcal{N}(\hat{\mathbf{Y}}, A  \mathbb V[\mathbf X | \mathbf Z]  A^\top +\delta ^2 \mathbf I),
\end{equation}
where $\hat{\mathbf{Y}} = \mathcal A(\hat{\mathbf X}_\theta(\mathbf Z))$, and $A\equiv\nabla\!_{\hat{\mathbf X}} \mathcal A$, i.e. the Jacobian of the non-linear function $\mathcal A$ w.r.t. $\hat{\mathbf X}_\theta(\mathbf Z)$. 
Hence, we approximate the posterior as
\begin{equation}
    \log p(\mathbf Y | \mathbf Z) = -\frac{1}{2}(\mathbf Y- \hat{\mathbf{Y}})^\top \Sigma_k^{-1}(\mathbf Y- \hat{\mathbf{Y}}) - \frac{1}{2} \log \left| 2 \pi \Sigma_k \right|,
\end{equation}
where $\Sigma_k = \beta(k)^2 A \nabla\!_{\mathbf{Z}} \hat{\mathbf{X}}\, A^\top + \delta^2 \mathbf{I}$. 
The gradient of $\log p(\mathbf Y | \mathbf Z)$ w.r.t. $\mathbf{Z}$ 
under the assumption that $\nabla\!_{\mathbf Z} \mathbb{V}[\mathbf{X}|\mathbf{Z}] \approx 0$ \citep{rozet2024learning} results in
\begin{equation} \label{eq:mmps_gradloglikelihood}
    \nabla\!_{\mathbf{Z}}\log q(\mathbf Y | \mathbf Z) = \nabla\!_{\mathbf{Z}} \hat{\mathbf{X}} A^\top \!\Sigma_k^{-1}(\mathbf Y- \hat{\mathbf{Y}}).
\end{equation}

Moreover, since the matrix $\Sigma_k$ is symmetric
positive definite, we obtain $\Sigma_k^{-1}(\mathbf Y- \hat{\mathbf{Y}})$ via conjugate gradient (CG) method, without computing $\Sigma_k^{-1}$ explicitly.
The CG method solves linear systems of the form $\mathbf{M} \mathbf{v} = \mathbf{b}$. Setting $\mathbf{M}=\Sigma_k$ and $\mathbf{b}=(\mathbf Y- \hat{\mathbf{Y}})$ we need to solve for $\mathbf{v}$.
This leads to 
\begin{align}
    (\mathbf Y- \hat{\mathbf{Y}}) &= 
    \left( \beta(k)^2 A \nabla\!_{\mathbf{Z}} \hat{\mathbf{X}} A^\top +\delta ^2 \mathbf I \right) \mathbf{v} \\
    \ &= \beta(k)^2 A \underbrace{\nabla\!_{\mathbf{Z}} \hat{\mathbf{X}} A^\top \mathbf{v}}_{\text{Jacobian-vector product}} +  \delta ^2 \mathbf{v}
\end{align}
After solving for $\mathbf{v}$ we can finally evaluate \cref{eq:mmps_gradloglikelihood}
\begin{equation}
    \nabla\!_{\mathbf{Z}}\log q(\mathbf Y | \mathbf Z) = \underbrace{ \nabla\!_{\mathbf{Z}} \hat{\mathbf{X}} A^\top \! \mathbf{v}}_{\text{Jacobian-vector product}}.
\end{equation}

\begin{algorithm*}[t]
\caption{WIND inverse problem solving}
\label{alg:wind_inverse}
\begin{algorithmic}
    \STATE {\bfseries Input:} Sequence length $T$, DDIM steps $N$, noise schedule $\{k_n\}$, stochasticity $\eta$
    \STATE {\bfseries Input:} Observation $\mathbf{Y}$, operator $\mathcal{A}$, task noise $\delta^2$, CG iterations $M$
    \FOR{$t = 1, \dots, T$}
        \STATE $\mathbf{z}^t \sim \mathcal{N}(\mathbf{0}, \mathbf{I})$ \COMMENT{Initialise all frames as pure noise}
    \ENDFOR
    \STATE $\mathbf{Z} \gets \{\mathbf{z}^1, \dots, \mathbf{z}^T\}$ \COMMENT{No clean context frame; all frames are noisy}
    \FOR{$n = N, N{-}1, \dots, 1$}
        \STATE $k \gets k_n$,\quad $k' \gets k_{n-1}$
        \STATE $\hat{\mathbf{X}} \gets \hat{\mathbf{X}}_\theta(\mathbf{Z})$ \COMMENT{Predict clean sequence given noised sequence $\mathbf{Z}$}
        \STATE $\mathbf{s}_\mathrm{lhd} \gets \mathrm{MMPS}(\hat{\mathbf{X}}_\theta(\mathbf{Z}),\, \mathbf{Y},\, \mathcal{A},\, k,\, \delta^2,\, M)$ \COMMENT{Likelihood score via \cref{alg:mmps}}
        \FOR{$t = 1, \dots, T$}
            \STATE $\tau \gets 1 - {\alpha(k)^2\,\beta(k')^2} / {\alpha(k')^2\,\beta(k)^2}$
            \STATE $\boldsymbol{\epsilon} \sim \mathcal{N}(\mathbf{0}, \mathbf{I})$
            \STATE $\mathbf s ^t \gets (\mathbf z^t -\alpha(k)\hat{\mathbf x} ^t)/\beta(k)^2$
            \STATE $\mathbf{s}^t \gets \mathbf{s}^t + \mathbf{s}_\mathrm{lhd}^t$ \COMMENT{Posterior guidance correction towards observation $\mathbf{Y}$}
            \STATE $\mathbf{z}^t \gets \alpha(k')\,\hat{\mathbf{x}}^t + \beta(k')\beta(k)\sqrt{1{-}\eta\tau}\,\mathbf s^t + \beta(k')\sqrt{\eta\tau}\,\boldsymbol{\epsilon}$ \COMMENT{DDIM update; $\eta{=}0$ deterministic, $\eta{=}1$ stochastic}
        \ENDFOR
    \ENDFOR
    \STATE \textbf{return} $\hat{\mathbf{X}}$ \COMMENT{Approximate posterior samples $p_\theta(\mathbf{X}|\mathbf{Y})$}
\end{algorithmic}
\end{algorithm*}

\begin{table*}[b]
\centering
\caption{Wall-clock runtime and peak memory for unguided and DAM-guided (MMPS) inference 
on the forecasting task.}
\label{tab:runtime}
\begin{tabular}{lrrrrrr}
\toprule
Run & DDIM Steps & CG Steps & Runtime [s] & Cost $\times$ vs Unguided & Peak Memory [GB] \\
\midrule
Unguided & 15 & - & 0.5573 & 1.00 & 9.59 \\
Unguided & 30 & - & 1.1133 & 1.00 & 9.59 \\
Guided   & 15 & 1 & 3.1930 & 5.74 & 14.82 \\
Guided   & 15 & 2 & 4.5189 & 8.12 & 14.82 \\
Guided   & 15 & 3 & 5.8466 & 10.51 & 14.82 \\
\bottomrule
\end{tabular}
\end{table*}

The complete MMPS procedure is summarised in \cref{alg:mmps}. 

\subsection{Inference Algorithms} \label{app:inference_algorithms}

We provide the complete inference procedures for the two operating modes
of \ourMethod: probabilistic forecasting (\cref{alg:wind_forecasting})
and inverse problem solving (\cref{alg:wind_inverse}). Both share
the same frozen pre-trained backbone $\hat{\mathbf{X}}_\theta$ and the
DDIM update derived in \cref{app:sampling_details}; they differ
only in (i) how the noised sequence $\mathbf{Z}$ is initialised and
(ii) whether the likelihood score from MMPS
(\cref{alg:mmps}) is injected into the per-frame score.

\paragraph{Forecasting mode (\cref{alg:wind_forecasting})}
Given one or more clean context frames $\mathbf{x}^0$, future frames are
initialised as pure Gaussian noise and combined with the clean
context into the sequence $\mathbf{Z}$, as also done in \citep{sestak2025lamslide}. 
Thanks to diffusion forcing, the
backbone natively accepts this mixed-noise input: clean frames are simply
treated as samples at noise level $k = 0$, while future frames are
progressively denoised over $N$ DDIM steps. When no observation operator
is supplied (\emph{unguided} branch), the procedure returns samples from
the learned prior $p_{\theta}(\mathbf{X} \mid \mathbf{x}^0)$, yielding an
ensemble forecast. When an additional observation $\mathbf{Y}$ and operator
$\mathcal{A}$ are supplied (\emph{guided} branch), the MMPS likelihood
score $\mathbf{s}_\mathrm{lhd}$ is added to the prior score at every DDIM
step, producing approximate posterior samples
$p_{\theta}(\mathbf{X} \mid \mathbf{x}^0, \mathbf{Y})$. 
This is the mode used when enforcing global dry air mass conservation
or in the the warmer-climate scenarios of \cref{sec:tasks}
\cref{fig:figure_2}a) shows a sketch of this mode.

\paragraph{Inverse problem mode (\cref{alg:wind_inverse})}
In contrast to forecasting, no clean context frame is provided. All $T$
frames of $\mathbf{Z}$ are initialised as pure Gaussian noise, and the
generation is steered entirely by the observation $\mathbf{Y}$ through
the MMPS likelihood score. This mode covers spatial and temporal
downscaling as well as sparse reconstruction.
Only the operator $\mathcal{A}$ and the target
measurement $\mathbf{Y}$ change between tasks, while the backbone, the
DDIM schedule, and the CG-based likelihood evaluation remain identical.

\begin{table*}[b]
\centering
\caption{CRPS comparison of WIND against state-of-the-art forecast models on WeatherBench2 \citep{rasp2024weatherbench} for the year 2020 across all variables at selected lead times (24h, 120h, 240h). Lower is better. WIND is trained at a coarser 1.5° resolution to learn a generalizable physical prior, whereas the other models are specialized for medium-range forecasting.}
\label{tab:scorecard}
\resizebox{\textwidth}{!}{%
\begin{tabular}{lcccccccccccc}
\hline
\textbf{Model} & \multicolumn{3}{c}{$T_{2m}\ [\mathrm{K}]$} & \multicolumn{3}{c}{$U_{10m}\ [\mathrm{m\,s^{-1}}]$} & \multicolumn{3}{c}{$V_{10m}\ [\mathrm{m\,s^{-1}}]$} & \multicolumn{3}{c}{$\mathrm{MSLP}\ [\mathrm{Pa}]$} \\
\cmidrule(lr){2-4} \cmidrule(lr){5-7} \cmidrule(lr){8-10} \cmidrule(lr){11-13}
 & 24h & 120h & 240h & 24h & 120h & 240h & 24h & 120h & 240h & 24h & 120h & 240h \\
\hline
WIND     & 0.286 & 0.624 & 0.921 & 0.511 & 1.229 & 1.688 & 0.542 & 1.290 & 1.775 & 38.810 & 167.379 & 274.268 \\
IFS ENS           & 0.396 & 0.610 & 0.911 & 0.415 & 1.000 & 1.602 & 0.424 & 1.036 & 1.674 & 32.056 & 118.719 & 247.034 \\
GenCast (oper.)   & 0.400 & 0.596 & 0.908 & 0.377 & 0.956 & 1.579 & 0.390 & 0.995 & 1.655 & 30.767 & 116.815 & 246.864 \\
GenCast           & 0.209 & 0.482 & 0.828 & 0.332 & 0.921 & 1.548 & 0.348 & 0.961 & 1.627 & 24.559 & 109.612 & 239.019 \\
ArchesWeatherGen  & 0.245 & 0.506 & 0.847 & 0.370 & 0.935 & 1.546 & 0.387 & 0.978 & 1.626 & 25.829 & 111.976 & 239.525 \\
Prob. Climatology & 1.123 & 1.123 & 1.125 & 1.873 & 1.875 & 1.875 & 1.919 & 1.918 & 1.917 & 321.175 & 321.869 & 322.199 \\
\hline
\end{tabular}%
}

\vspace{0.5em}

\resizebox{\textwidth}{!}{%
\begin{tabular}{lcccccccccccc}
\hline
\textbf{Model} & \multicolumn{3}{c}{$T_{500}\ [\mathrm{K}]$} & \multicolumn{3}{c}{$Q_{500}\ [\mathrm{g\,kg^{-1}}]$} & \multicolumn{3}{c}{$Q_{850}\ [\mathrm{g\,kg^{-1}}]$} & \multicolumn{3}{c}{$\text{TP}_{6h}\ [\mathrm{mm}]$} \\
\cmidrule(lr){2-4} \cmidrule(lr){5-7} \cmidrule(lr){8-10} \cmidrule(lr){11-13}
 & 24h & 120h & 240h & 24h & 120h & 240h & 24h & 120h & 240h & 24h & 120h & 240h \\
\hline
WIND     & 0.362 & 0.884 & 1.316 & 165.1 & 285.3 & 357.5 & 369.6 & 659.4 & 836.5 & 0.300 & 0.441 & 0.501 \\
IFS ENS           & 0.243 & 0.672 & 1.206 & 127.3 & 257.4 & 346.0 & 339.8 & 614.1 & 818.9 & 0.253 & 0.416 & 0.496 \\
GenCast (oper.)   & 0.240 & 0.647 & 1.189 & 120.2 & 239.9 & 333.3 & 327.8 & 578.7 & 794.8 & —     & —     & —     \\
GenCast           & 0.213 & 0.627 & 1.163 & 111.5 & 236.4 & 328.8 & 274.7 & 563.5 & 781.9 & —     & —     & —     \\
ArchesWeatherGen  & 0.238 & 0.635 & 1.165 & 120.4 & 234.7 & 327.5 & 294.1 & 560.1 & 779.6 & —     & —     & —     \\
Prob. Climatology & 1.558 & 1.558 & 1.558 & 401.5 & 401.2 & 401.1 & 926.5 & 925.8 & 924.9 & 0.524 & 0.523 & 0.523 \\
\hline
\end{tabular}%
}
\end{table*}

\subsection{Computational Cost}
\cref{tab:runtime} reports wall-clock runtime, cost relative to unguided sampling, and peak 
memory usage, averaged over 10 runs after 3 burn-in runs on a NVIDIA RTX PRO 6000. 
We benchmark the forecasting task, where we generate 4 future frames conditioned on 1 clean context frame. Guided runs additionally enforce 
dry air mass (DAM) conservation via MMPS at every denoising step. Unguided sampling scales 
approximately linearly with DDIM steps, running in $0.56$\,s for 15 steps and $1.11$\,s for 30 
steps, with a peak memory footprint of $9.6$\,GB in both cases. Guided sampling via MMPS 
introduces additional cost due to the conjugate gradient (CG) solve at each denoising step: with 
15 DDIM steps, runtime scales approximately linearly with CG iterations, from $3.2$\,s ($1$ CG 
step, $5.7\times$ overhead) to $5.8$\,s ($3$ CG steps, $10.5\times$ overhead), with peak memory 
increasing to $14.8$\,GB. 

\section{Downstream Task Details} 

\subsection{Forecasting} \label{app:forecasting_details}

\begin{figure*}[t]
    \centering
    \includegraphics[width=0.99\linewidth]{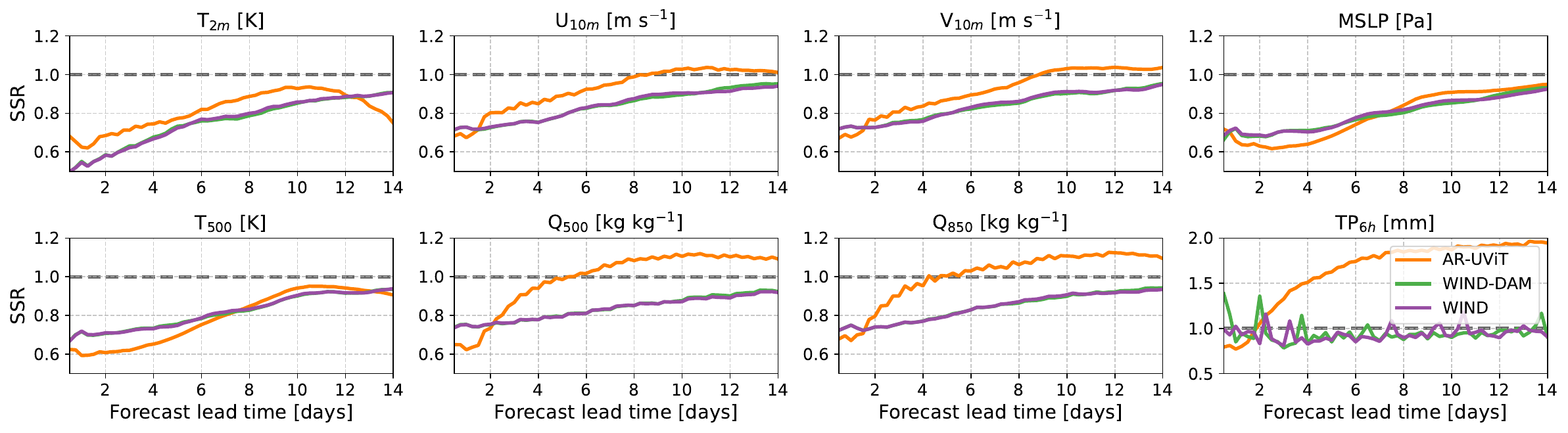}
    \caption{\textbf{Spread-skill ratio.} 
    We assess the reliability of the probabilistic forecast using the spread-skill ratio (SSR), where a value of 1.0 (dashed line) indicates perfect calibration. 
    \ourMethod and its physically constrained variant \ourMethod-DAM start slightly over-confident (SSR $<1$) but steadily approach ideal calibration over the 14-day horizon without overshooting. 
    In contrast, the autoregressive AR-UViT baseline exhibits rapidly increasing spread, drifting into under-confidence (SSR $>1$) for variables like specific humidity ($Q_{500}$) and total precipitation (${TP}_{6h}$). 
    The significant overlap between the purple and green lines confirms that enforcing the dry air mass constraint preserves the probabilistic calibration of the ensemble.}
    \label{fig:forecasting_ssr}
\end{figure*}

\begin{figure*}[b]
    \centering
    \includegraphics[width=0.95\linewidth]{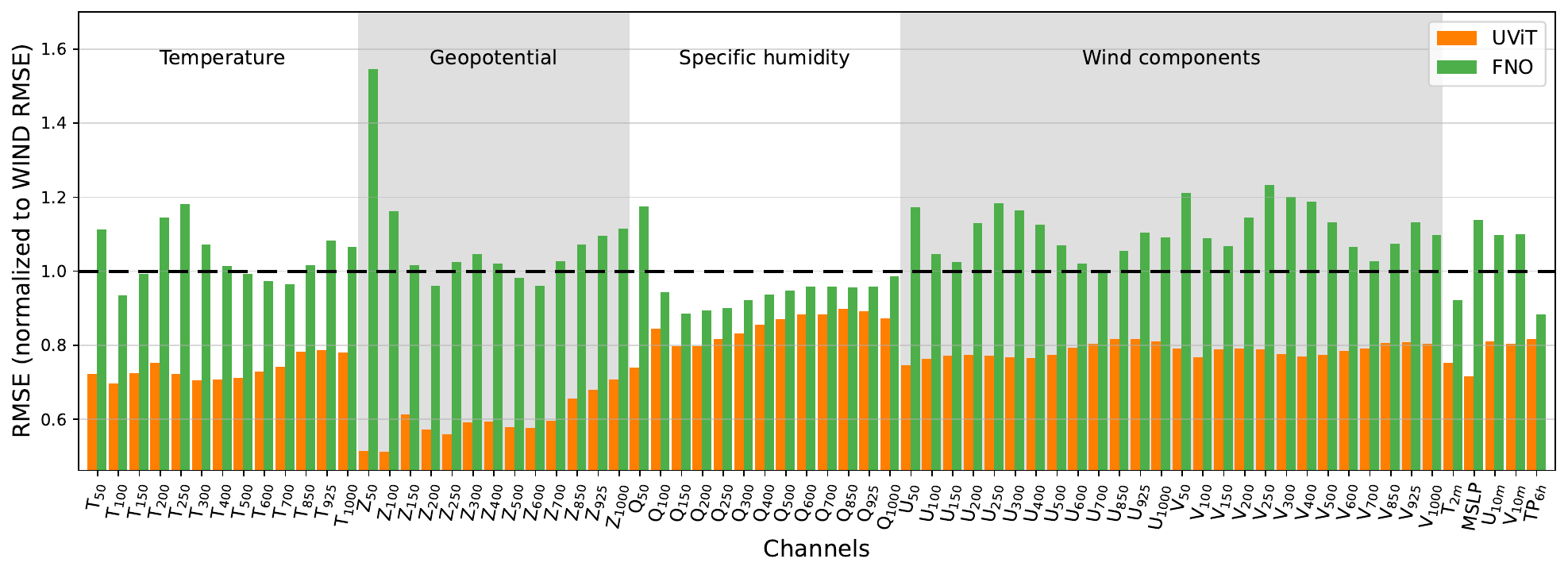}
    \caption{\textbf{RMSE comparison of spatial downscaling.} We compare the RMSE of the baselines relative to our method (dashed line at 1.0) to account for different scales in different fields. While the specialized UViT achieves lower RMSE by optimizing for the mean, \ourMethod outperforms the FNO on several variables and remains competitive on others, despite not being trained on the task.} 
    \label{fig:spatial_downscaling_rmse_normalized}
\end{figure*}

\paragraph{Baselines} \label{gencast_baseline}
To simulate the GenCast architecture, we train our model with a sequence length of $T=3$. In this configuration, we condition on two clean frames and denoise only the final frame, setting $\mathbf{k} = \{0, 0, k\}$. This effectively models the transition probability $p(\mathbf{x}^{t} | \mathbf{x}^{t-2}, \mathbf{x}^{t-1})$.
Additionally, to evaluate the advantages of diffusion forcing for long-range rollouts, we also trained a variant called UViT-5 ($T=5$). In this setup, we provide a single clean frame and simultaneously denoise the subsequent four frames at a uniform noise level, such that $\mathbf{k} = \{0, k, k, k, k\}$.

\paragraph{Evaluation metrics}
Due to the highly chaotic dynamics of the atmosphere, probabilistic forecasts accompanied by uncertainty quantification are key for evaluating weather forecast models. The spread-skill ratio (SSR) \citep{fortin2014should} and continuous ranked probability score (CRPS) \citep{gneiting2007strictly} are the standard ensemble metrics to assess the forecast performance. The SSR tests the models calibration by comparing the ensemble's spread (standard deviation) to the root-mean-square error of the ensemble mean, where a ratio close to 1 indicates that the forecast uncertainty accurately reflects the actual error.
The spread is defined as:
\begin{equation}
    \textrm{Spread} = \sqrt{\frac{1}{HW} \sum_{h,w} a_{h,w} \frac{1}{M-1} \sum_{m=1}^M ( \hat{\mathbf{x}}_{m,h,w} - \bar{\mathbf{x}}_{h,w} )^2} ,
\end{equation}
with $\bar{\mathbf{x}}_{h,w} = \frac{1}{M} \sum_{m=1}^M \hat{\mathbf{x}}_{m,h,w}$ as the ensemble mean of the predictions.
The skill is defined as:
\begin{equation}
    \textrm{Skill} = \sqrt{\frac{1}{HW} \sum_{h,w} a_{h,w} ( \bar{\mathbf{x}}_{h,w} - \mathbf{x}_{h,w} )^2} .
\end{equation}
Combining both results in the SSR: 
\begin{equation}
    \textrm{SSR} = \sqrt{\frac{M+1}{M}} \frac{\textrm{Spread}}{\textrm{Skill}} .
\end{equation}

The CRPS measures the overall accuracy of a probabilistic forecast by quantifying the integrated squared difference between the forecast's cumulative distribution function and the observed step function, penalizing both bias and lack of sharpness. The CRPS acts as a mean absolute error for probabilistic forecasts, measuring the distance between the entire range of predicted forecasts and the single ground truth observation. 
We use the following CRPS definition:
\begin{align}
    \textrm{CRPS} &= \frac{1}{HW} \sum_{h,w} a_{h,w} \biggl[ \frac{1}{M} \sum_{m=1}^M \lvert \hat{\mathbf{x}}_{m,h,w} - \mathbf{x}_{h,w} \rvert \notag \\
    & - \frac{1}{2M(M-1)} \sum_{m=1}^M \sum_{m'=1}^M \lvert \hat{\mathbf{x}}_{m,h,w} - \hat{\mathbf{x}}_{m',h,w} \rvert
    \biggr].
\end{align}

\begin{figure*}
    \centering
    \includegraphics[width=0.99\linewidth]{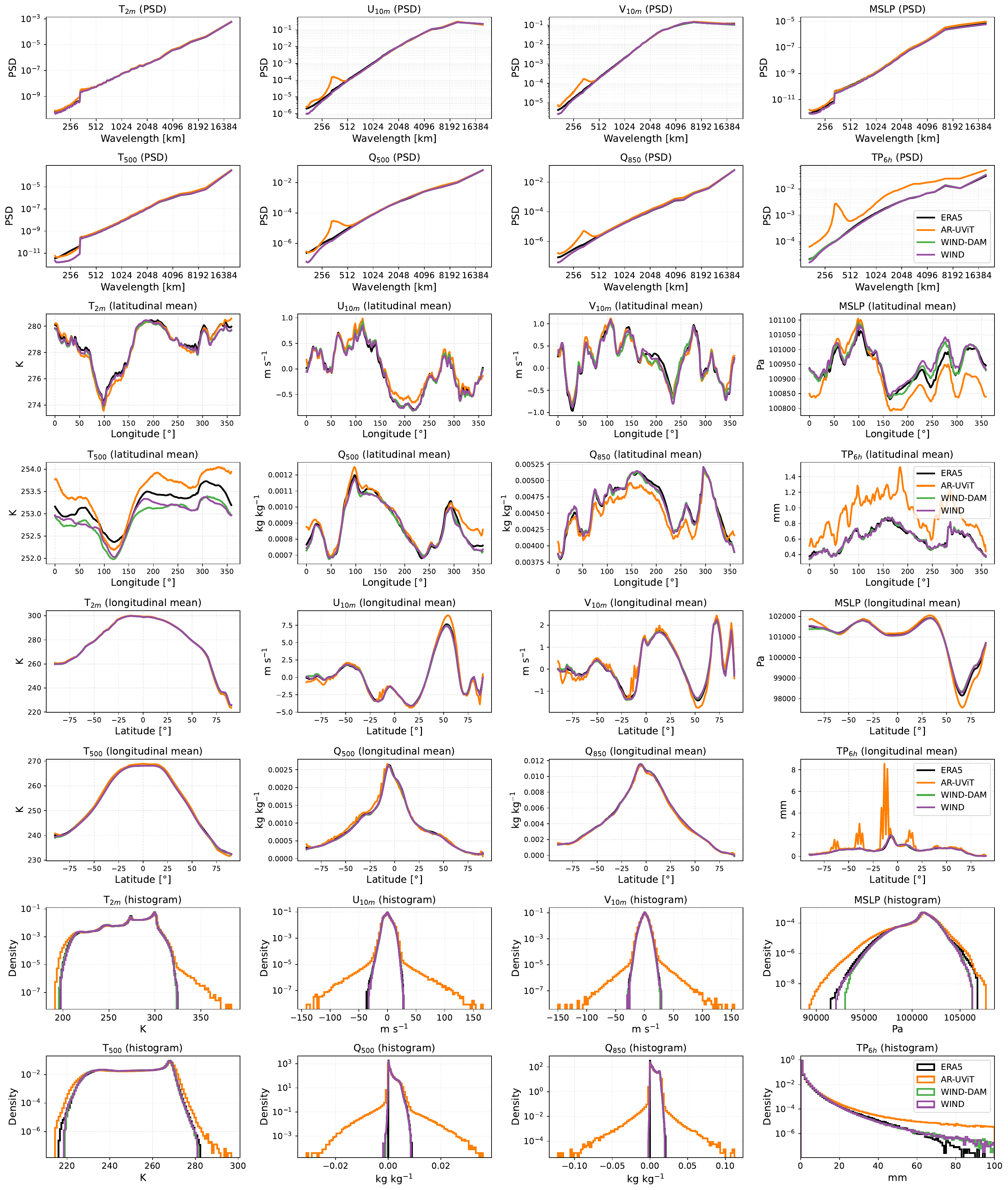}
    \caption{\textbf{Spectrum and distribution for forecasting.} Top rows: WIND accurately models the forecast spectrum; in contrast, AR-UViT struggles with high frequencies and precipitation modeling. Middle rows: AR-UViT overestimates precipitation, with notable artifacts near the equator. Bottom rows: AR-UViT histograms shows increasing values in all fields from repeated autoregressive steps, whereas WIND accurately tracks the ERA5 ground truth.}
    \label{fig:forecasting_psd_latlon_mean_hist}
\end{figure*}

\paragraph{Extended results}
In addition to the CRPS results in the main paper, we additionally visualize the spectrum and the distribution for the forecasting in \cref{fig:forecasting_psd_latlon_mean_hist}. 
Additionally we compare WIND against state-of-the-art forecasts models on WeatherBench2 in \cref{tab:scorecard}.

\subsection{Spatial Downscaling} \label{app:spatial_downscaling}

\begin{figure*}[t]
    \centering
    \includegraphics[width=0.95\linewidth]{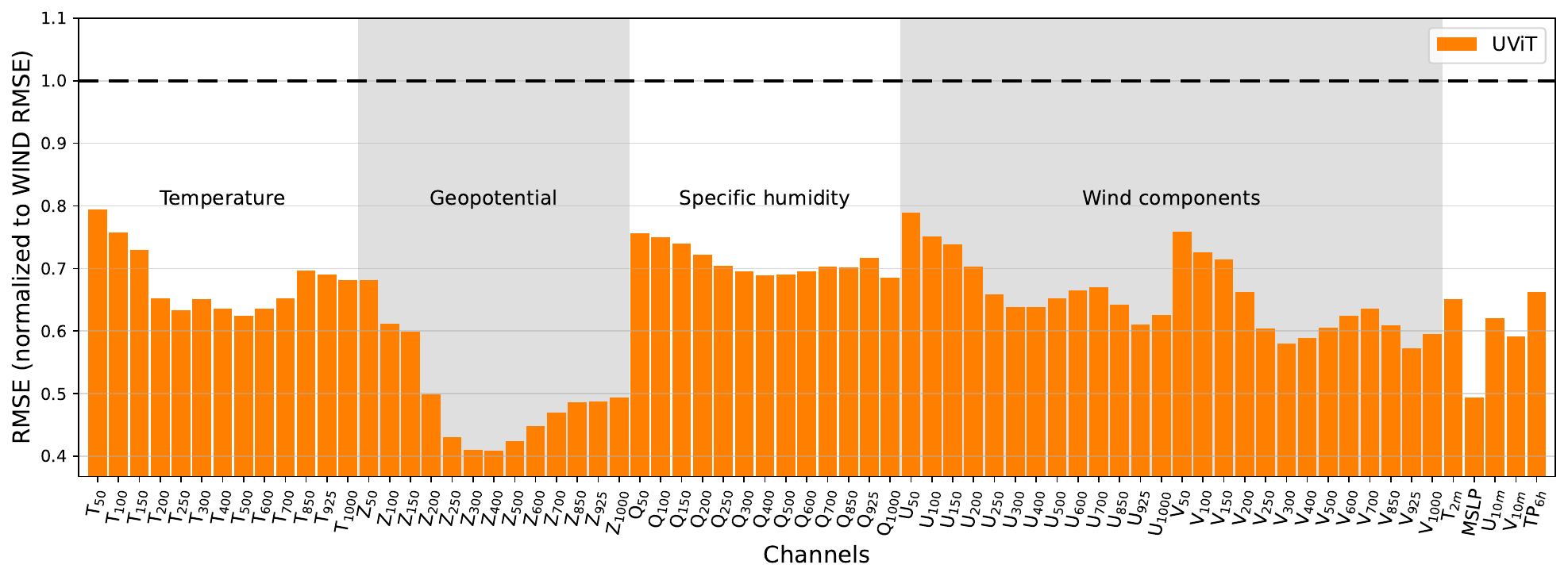}
    \caption{\textbf{RMSE comparison of temporal downscaling.} We compare the RMSE of the baselines relative to our method (dashed line at 1.0) to account for different scales in different fields. The specialized UViT achieves lower RMSE than \ourMethod. }
    \label{fig:temporal_downscaling_rmse_normalized}
\end{figure*}

\paragraph{Formulation} \label{app:spatial_formulation}
We aim to recover a high-resolution sequence $\mathbf{X} \in \mathbb{R}^{T \times C \times H \times W}$ given a low-resolution observation sequence $\mathbf{Y} \in \mathbb{R}^{T \times C \times (H/s) \times (W/s)}$. Here, $T=5$ denotes the sequence length and $s \in \mathbb{N}$ is the downscaling factor.
In our experiment, the target $\mathbf{X}$ consists of $1.5^\circ$ ERA5 fields, while the condition $\mathbf{Y}$ is the corresponding $6^\circ$ low-resolution sequence ($s=4$). The relationship is defined by the forward operator $\mathcal{A}$ applied frame-wise for $t=1 \dots T$:
\begin{equation}
    \mathbf{y}_t = \mathcal{A}(\mathbf{x}_t) = \text{AvgPool}_{s \times s}(\mathbf{x}_t).
\end{equation}

\begin{table}[h]
\centering
\caption{\textbf{Quantitative comparison for spatial downscaling.} We report the absolute RMSE averaged over all pressure levels. While the specialized UViT baseline achieves the lowest RMSE, \ourMethod outperforms the specialized FNO without any task-specific training.}
\label{tab:spatial_downscaling_rmse_comparison}
\begin{tabular}{lrrr}
\toprule
Variable & WIND & UViT & FNO \\
\midrule
Temperature (3D) & \underline{0.63} & \textbf{0.47} & 0.66 \\
Geopotential (3D) & \underline{45.17} & \textbf{25.86} & 52.44 \\
Specific humidity (3D) & 0.0005 & 0.0005 & 0.0005 \\
U-Wind (3D) & \underline{1.89} & \textbf{1.47} & 2.08 \\
V-Wind (3D) & \underline{1.76} & \textbf{1.38} & 2.01 \\
\midrule
2m temperature & 0.76 & \textbf{0.57} & \underline{0.70} \\
MSLP & \underline{42.68} & \textbf{30.58} & 48.53 \\
10m U-Wind & \underline{0.93} & \textbf{0.76} & 1.02 \\
10m V-Wind & \underline{0.95} & \textbf{0.76} & 1.04 \\
Precipitation & 1.77 & \textbf{1.43} & \underline{1.55} \\
\bottomrule
\end{tabular}
\end{table}

\paragraph{Baselines}
We benchmark our approach against specialized architectures optimized for spatial downscaling. Specifically, we train the FNO and UViT described in \cref{app:architecture_hp} as deterministic mappings between the input and target resolutions. To account for the downscaling factor of $s=4$ while maintaining the models' requirement for consistent input-output dimensions, the low-resolution inputs are projected back to the high-resolution grid using nearest-neighbor interpolation. All baseline models operate on individual frames independently, representing a temporal constraint of $T=1$.

\begin{figure*}
    \centering
    \includegraphics[width=0.99\linewidth]{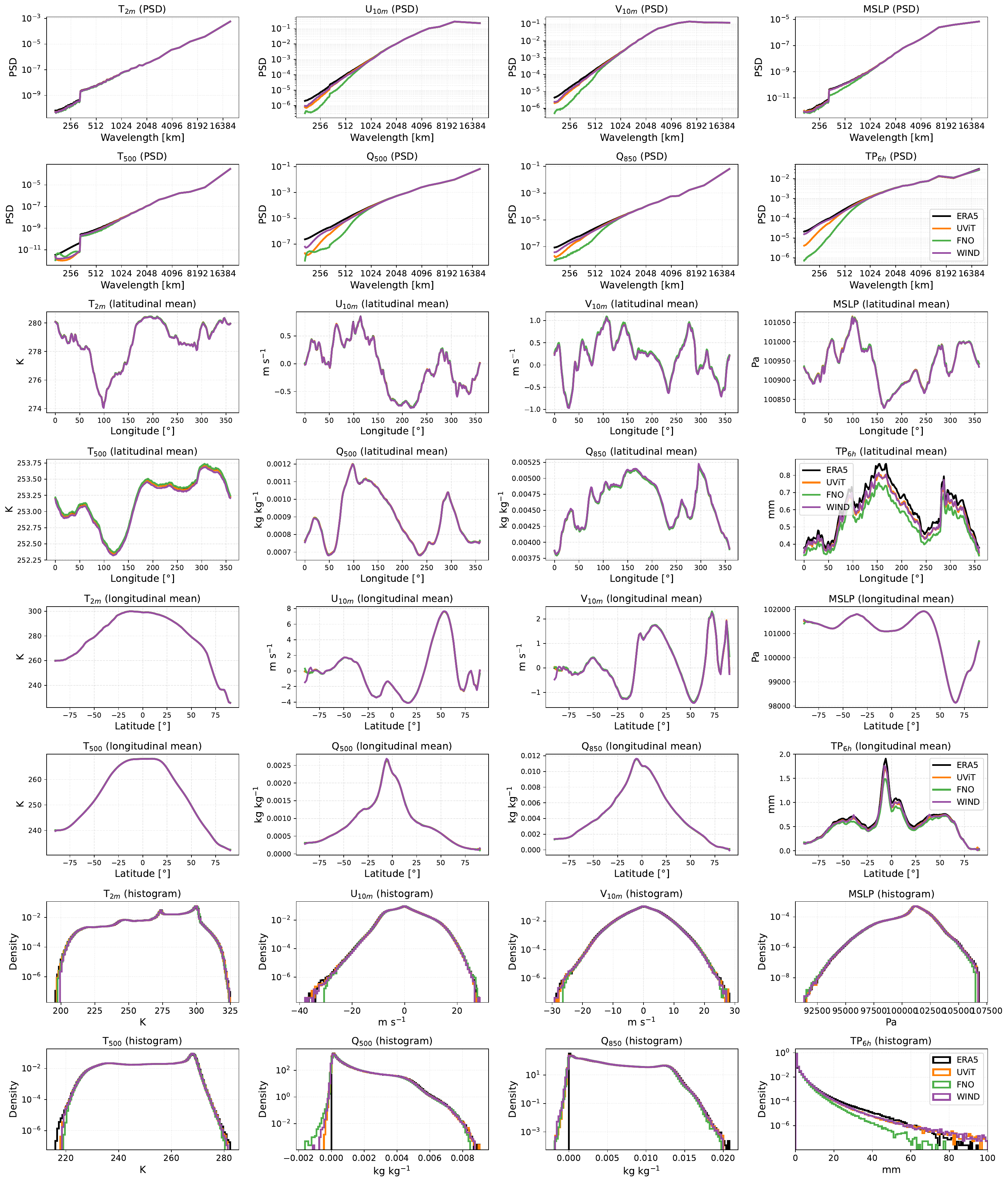}
    \caption{\textbf{Spectrum and distribution for spatially downscaled fields.} Top rows: We compare the PSD of the ERA5 ground truth, deterministic FNO, UViT and \ourMethod. \ourMethod closely tracks the energy spectrum of ERA5 across all scales, preserving high-frequency details. In contrast, the deterministic FNO baseline exhibit spectral drop-off at high frequencies. Middle rows: \ourMethod and UViT perform on par at reproducing the latitudinal and longitudinal means. FNO is worse for precipitation. Bottom rows: The histograms confirm that observation.}
    \label{fig:spatial_downscaling_psd_latlon_mean_hist}
\end{figure*}

\paragraph{Extended results} \label{app:spatial_results}
In addition to the power spectrum shown in the main text, the histograms in \cref{fig:spatial_downscaling_psd_latlon_mean_hist} confirm that \ourMethod preserves the statistical properties of the atmosphere. For heavy-tailed variables like 6-hour precipitation ($\text{TP}_{6h}$) and specific humidity ($Q_{500}$ and $Q_{850}$), our model successfully reproduces the distribution of the high-resolution data, whereas deterministic baselines often under-predict extremes.

We also tested how well the predictions align with the low-resolution condition. After re-applying the operator $\mathcal{A}$, we found a pearson correlation of 0.96 for \ourMethod, matching the specialized baselines. \cref{tab:spatial_downscaling_rmse_comparison} provides the full RMSE breakdown while \cref{fig:spatial_downscaling_rmse_normalized} compares the normalized RMSEs of our baselines and \ourMethod. While pixel-wise metrics favor the specialized UViT, \ourMethod remains competitive with the FNO baseline despite not being trained on the task.

\subsection{Temporal Downscaling} \label{app:temporal_details}

\paragraph{Formulation}
Temporal downscaling aims to recover a high-frequency sequence $\mathbf{X} \in \mathbb{R}^{T \times C \times H \times W}$ given a temporally aggregated observation $\mathbf{Y} \in \mathbb{R}^{1 \times C \times H \times W}$ (e.g. a daily mean). For our 6-hourly data, $T=4$ corresponds to the snapshots within a 24-hour window.
The relationship is defined by the forward operator $\mathcal{A}$, which aggregates frames over the temporal dimension, effectively smoothing them:
\begin{equation}
    \mathbf{Y} = \mathcal{A}(\mathbf{X}) = 
    \frac{1}{T} \sum_{k=1}^{T} \mathbf{x}_k .
\end{equation}

Our model is trained on a window of $K=5$ frames. To perform temporal downscaling, we generate a sequence $\mathbf{X}$ where the first $T=4$ frames are constrained to match the daily observation $\mathbf{Y}$. The fifth frame is unconstrained, demonstrating the model's flexibility to handle cases where the window size is not perfectly aligned with a downstream task. We sample from the posterior $p_\theta(\mathbf{X} | \mathbf{Y})$ using MMPS guidance, ensuring that the temporal aggregation of the generated sequence is consistent with the daily ground truth. In principle, the target resolution is constrained only by the window size. We evaluate the ability to recover high-frequency patterns lost during the temporal averaging, alongside the ability to still align with the daily average field. We temporally downscale every day in 2021 from the ERA5 ground truth and compare to the 6-hourly ground truth.

\paragraph{Baselines} 
For temporal downscaling, we employ a UViT baseline with a temporal horizon of $T=4$. The model receives the daily mean—repeated across all four temporal input slots and is trained via MSE loss to regress the original 6-hourly sequences.

\begin{figure}[h]
    \centering
    \includegraphics[width=0.95\linewidth]{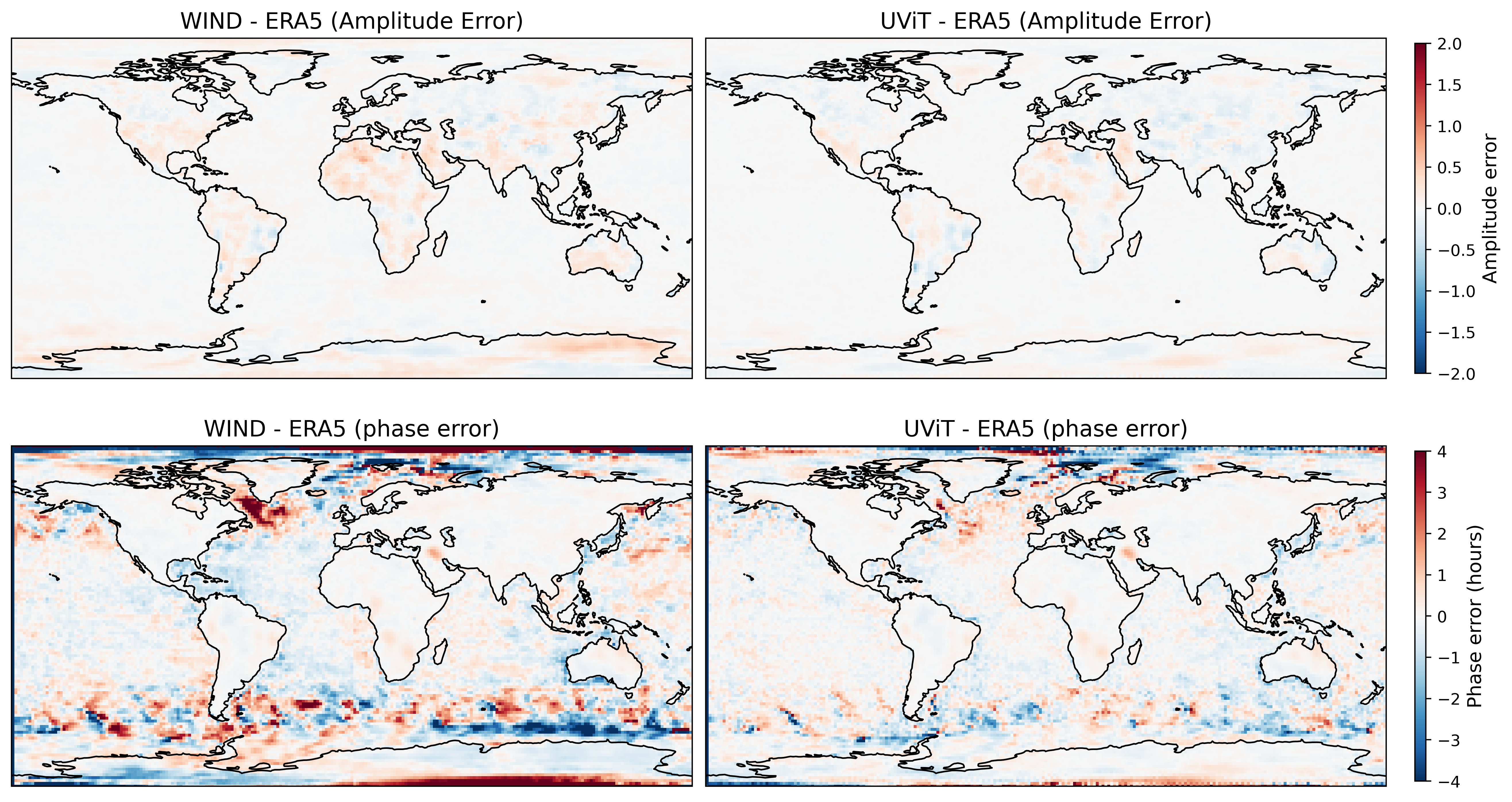}
     \caption{\textbf{Validation of sub-daily dynamics via diurnal harmonic analysis.} We evaluate the physical fidelity of the generated 6-hourly fields by decomposing the diurnal cycle into amplitude and phase for 2m Temperature. The plots show the pixel-wise error relative to ERA5 ground truth for \ourMethod and UViT. (Top Row) amplitude bias: the daily mean baseline (right) exhibits a massive negative bias, confirming the significant variance lost by averaging. Both diffusion models (left/middle) exhibit near-zero bias, demonstrating accurate recovery of the diurnal range. (Bottom Row) phase lag: measuring the shift in peak time relative to ERA5. Over land, both models show negligible phase error, correctly capturing the thermal lag of peak temperature. Note that phase noise over oceans is expected due to the negligible diurnal amplitude in those regions.} 
    \label{fig:diurnal_cycle}
\end{figure}

\paragraph{Extended results} \label{app:temporal_results}

\begin{figure*}[t!]
    \centering
    \includegraphics[width=0.99\linewidth]{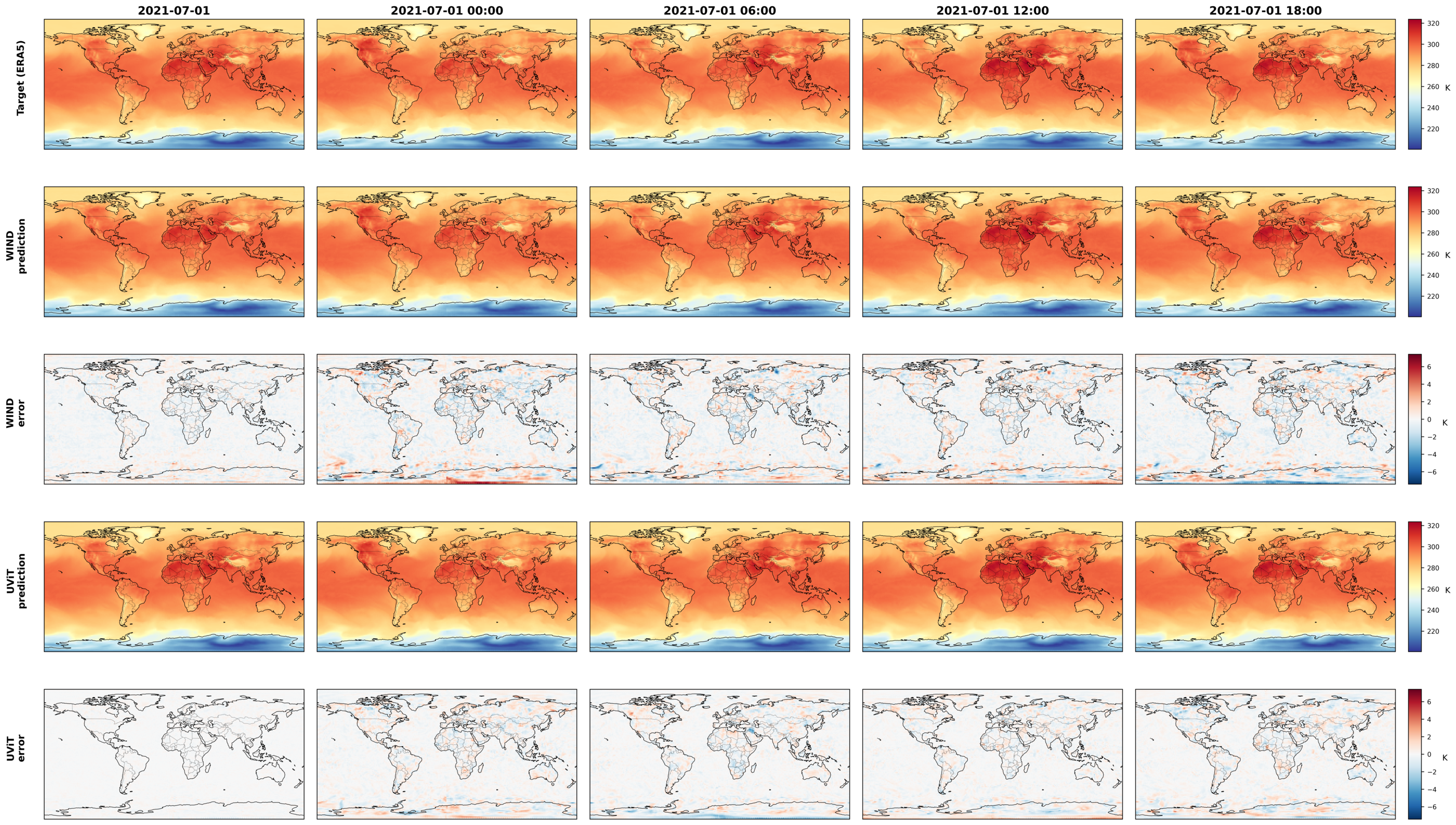}
    \caption{\textbf{Qualitative comparison of temporal downscaling for 2m temperature.} The first column displays the daily average, while the subsequent four columns show the 6-hourly high-frequency sequences. The top row shows the ERA5 target on July 1, 2021. The second and fourth rows show predictions from \ourMethod and the specialized UViT baseline, respectively, with their corresponding pixel-wise error maps shown in the third and fifth rows.}
    \label{fig:temporal_downscaling_2m_temperature_comparison}
\end{figure*}

Quantitatively, the specialized UViT baseline achieves lower RMSE, as shown in \cref{tab:rmse_comparison_temp_DS} and \cref{fig:temporal_downscaling_rmse_normalized}. Analogous to the spatial case, slight temporal phase shifts are heavily penalized by pixel-wise metrics like RMSE, even when the generated weather is physically valid. 
\cref{fig:temporal_downscaling_2m_temperature_comparison} compares \ourMethod and UViT qualitatively; both models perform on par. To further validate physical fidelity, \cref{fig:diurnal_cycle} decomposes the diurnal cycle into harmonic amplitude and phase. \ourMethod demonstrates near-zero amplitude bias and accurate phase locking. This confirms that, despite higher RMSE, the model captures the thermodynamics and thermal inertia correctly. 

\begin{table}[htbp]
\centering
\caption{\textbf{Quantitative comparison for temporal downscaling.} We report the absolute RMSE averaged over all pressure levels for \ourMethod and a task specific diffusion baseline (UViT). The conditional UViT baseline achieves a lower RMSE.}
\label{tab:rmse_comparison_temp_DS}

\begin{tabular}{lrr}
\toprule
Variable & WIND & UViT \\
\midrule
Temperature (3D) & 0.79 & \textbf{0.55} \\
Geopotential (3D) & 84.47 & \textbf{42.06} \\
Specific Humidity (3D) & 0.0004 & \textbf{0.0003} \\
U-Wind (3D) & 2.36 & \textbf{1.61} \\
V-Wind (3D) & 2.66 & \textbf{1.70} \\
\midrule
2m Temperature & 0.71 & \textbf{0.46} \\
MSLP & 83.91 & \textbf{40.98} \\
10m U-Wind & 1.09 & \textbf{0.67} \\
10m V-Wind & 1.22 & \textbf{0.72} \\
Precipitation & 1.68 & \textbf{1.10} \\
\bottomrule
\end{tabular}
\end{table}

\subsection{Sparse Reconstruction} \label{app:sparse_details}

\paragraph{Background}

Satellite measurements are inherently limited by orbital characteristics that create spatial discontinuities. Consequently, the ability to reconstruct global atmospheric states from sparse observations is critical for both modern analysis and extending datasets into historical eras lacking satellite coverage. Diffusion models have emerged as powerful tools for such reconstruction tasks in image processing \citep{saharia2022palette}, PDEs \citep{amoros2026guiding}, and weather \citep{li2024learning, kishikawa2025conditional, qian2026generative}. Recently, \citet{li2024learning}, proposed a specialized framework (S³GM) to perform sparse reconstruction at inference time using a custom conditional SDE solver. In this work, we demonstrate that the physics prior of our pre-trained diffusion model can solve the sparse reconstruction task purely at inference time using MMPS. This approach eliminates the need to train specialized conditional models specifically for sparse reconstruction tasks and complex sampler modifications. We compare our approach against traditional baselines such as Kriging \citep{cressie1990origins}, a Gaussian Process, based interpolation method widely used in geosciences. Unlike traditional approaches, MMPS is completely independent of the sensor and does not need to be tuned for different sensor types, compared to traditionally statistical or conditional-training-based methods.

\textbf{Formulation}
We frame sparse reconstruction as the recovery of the full global atmospheric state $\mathbf{X}$ from a set of sparse, point-wise observations $\mathbf{Y}$. Referring to the general inverse formulation in \cref{sec:approach}, the forward operator $\mathcal{A}$ for this task is defined as a binary masking operation:
\begin{equation}
    \mathbf{Y} = \mathcal{A}(\mathbf{X}) = \mathbf{M} \odot \mathbf{X}
\end{equation}
where $\mathbf{M}$ is a binary mask representing the spatial locations of sensors (e.g. weather stations or satellite tracks).

\paragraph{Baselines} 
For spatial reconstruction, we train a UViT ($T=1$) to recover the full state $\mathbf{X}$ from sparse observations $\mathbf{Y}$. The model receives the masked observations concatenated with the binary mask as an additional input channel. We employ a dynamic masking strategy, sampling sparsity levels between 1\% and 10\% with randomized spatial patterns. The architecture is optimized via MSE loss to reconstruct the complete field from these partial observations.

\begin{table}[h]
\centering
\caption{\textbf{Comparing absolute RMSEs for sparse reconstruction with sparsity $1\%$.} We compare inference-only method \ourMethod against a Gaussian Process baseline (Kriging) and a specialized conditional diffusion model (UViT). For the atmospheric variables we averaged the RMSE values over all pressure levels. \ourMethod outperforms the baselines on most of the variables.}
\label{tab:rmse_comparison_sparse_1}
\begin{tabular}{lrr}
\toprule
Variable & WIND & UViT \\
\midrule
Temperature (3D) & \textbf{0.65} & 0.68 \\
Geopotential (3D) & \textbf{48.64} & 80.97 \\
Specific Humidity (3D) & \textbf{0.0006} & 0.0007 \\
U-Wind (3D) & \textbf{1.84} & 1.87 \\
V-Wind (3D) & 1.85 & \textbf{1.84} \\
\midrule
2m Temperature & \textbf{0.83} & 0.86 \\
MSLP & \textbf{47.12} & 51.19 \\
10m U-Wind & \textbf{0.95} & 0.99 \\
10m V-Wind & \textbf{1.00} & 1.01 \\
Precipitation & 0.0017 & \textbf{0.0016} \\
\bottomrule
\end{tabular}
\end{table}

\subsection{Enforcing Global Dry Air Mass Conservation} \label{app:conservation_details}

\paragraph{Motivation}
A major limitation of purely data-driven AI forecasting models is that they often become unstable for longer rollouts. While these instabilities come partly from architectural choices, they are also rooted in the inability of AI models to accurately obey the underlying physical conservation laws. 
To address this, recent literature has tried to enforce physical quantities directly. The two leading approaches are modifying the loss function via soft constraints \citep{verma2024climode} or adding specific neural network layers as hard constraints to strictly enforce global conservation of these quantities \citep{sha2025improving,harder2023hard,watt2025ace2}. 
Most AI-based weather models do not enforce conservation laws, partly because they are trained on ERA5 reanalysis data \citep{hersbach2020era5}, which itself does not strictly conserve mass and energy due to the underlying data assimilation process \citep{tootoonchi2025revisiting}. Recent work by \citet{sha2025improving} demonstrated that enforcing global conservation of energy, moisture budget and DAM improves forecast performance particularly for precipitation, while reducing the drizzle bias (the tendency of AI models to predict light rain everywhere).

\paragraph{Formulation} We enforce constant dry air mass via  $\mathcal{A}(\mathbf{X}) = \text{f}_{\text{DAM}}( \mathbf{x}^t) = C_{\text{DAM}}$.
The function $\text{f}_{\text{DAM}}$ is derived as: 
\begin{align} \label{eq:dam_computation}
    p_{\text{sfc}} &= p_{\text{MSLP}} \exp \left( -\frac{\Phi_{\text{sfc}}}{R_d T_{\text{2m}}} \right) \nonumber \\
    \text{TWP} &= \frac{1}{g} \int_{p_{\text{top}}}^{p_{\text{sfc}}} Q(p) \, dp \nonumber \\
    m_{\text{dry}} &= \frac{p_{\text{sfc}}}{g} - \text{TWP} \nonumber \\
    \text{f}_{\text{DAM}}(\mathbf{x}^t) &= \sum_{h,w} a_{h,w} m_{\text{dry}}(h,w)
\end{align}
where $p_{\text{sfc}}$ is the derived surface pressure and $p_{\text{MSLP}}$ is the mean sea level pressure. The term $\Phi_{\text{sfc}}$ represents the surface geopotential, $R_d$ is the gas constant for dry air, and $T_{\text{2m}}$ is the 2m temperature. For the vertical moisture integration, $g$ denotes gravitational acceleration, $Q(p)$ is the specific humidity at pressure level $p$, and $\text{TWP}$ is the total water path. Finally, $a_{h,w}$ represents the area weighting. In \cref{fig:dam_time}, we guide the rollout using the numerical DAM value of the first clean frame, denoted as $C_{\text{DAM}}$, which is used to initialize the rollout.

\subsection{Generating Weather in a Warmer Climate} \label{app:storyline_details}
We compare an unguided ensemble run starting from the historical initial condition to two warming runs, one guided and one evolving freely. In both warm runs we perturb the initial state by adding $+2\text{K}$ to all temperature channels and scaling the specific humidity channels by a factor of $1.07^{\Delta T=2K}$ (based on Clausius-Clapeyron scaling), leaving dynamic variables unchanged. 

In the guided run we enforce the thermodynamic anomaly throughout the generation process by framing it as an inverse problem. We define the $\mathcal{A}$ operator as the spatial average over the temperature and specific humidity fields. The target $\mathbf{y}$ contains the spatial average for all thermodynamic variables for each frame. The average is computed over the historical ground truth. We increase the temperature means $\mathbf{y_{\text{T}}}$ by +2K and increase the specific humidity means $\mathbf{y_{\text{Q}}}$ by +14\%. To address the numerical scale disparity between temperature ($\sim 10^2$) and humidity ($\sim 10^{-3}$), we weight the respective channels based on the inverse magnitude of the perturbations to ensure balanced gradient contributions. Inference is performed using 10 steps with $\eta = 1$, 10 conjugate gradient steps and noise variance $\delta^2 = 1e^{-3}$. 

\begin{figure}[t!]
    \centering
    \includegraphics[width=0.99\linewidth]{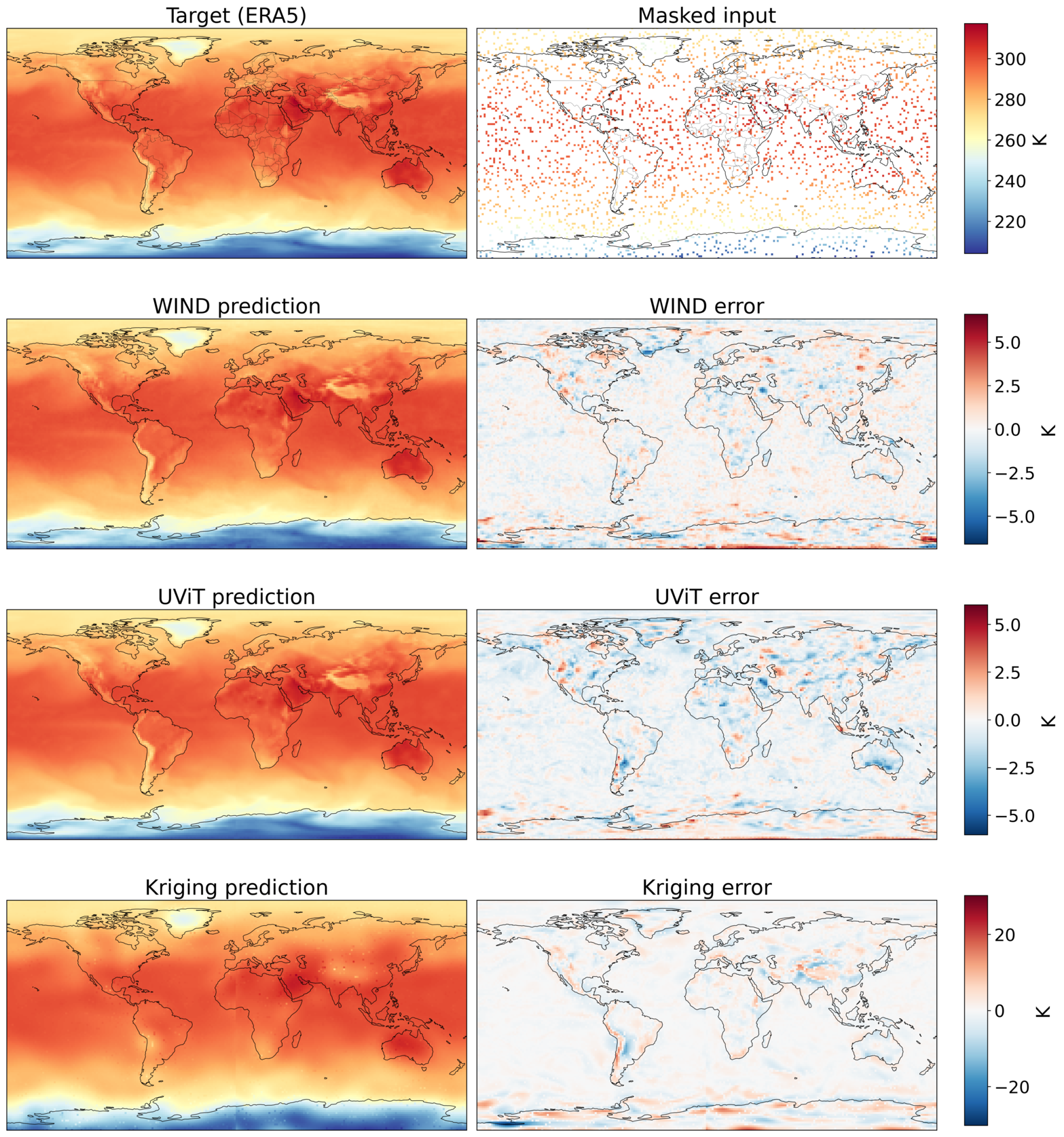}
    \caption{\textbf{Qualitative comparison of sparse reconstruction.} 
    \textbf{Left:} ERA5 ground truth (2m Temperature) and \textbf{10\%} sparse input mask. 
    \textbf{Center:} Predictions from \ourMethod, UViT, and Kriging. 
    \textbf{Right:} Prediction error relative to ground truth. 
    While \ourMethod and UViT recover physically coherent fields with realistic gradients, Kriging yields overly smooth interpolations that miss fine-grained patterns.} 
    \label{fig:sparse_reconstruction_example_0_1}
\end{figure}

\begin{figure}[t!]
    \centering
    \includegraphics[width=0.99\linewidth]{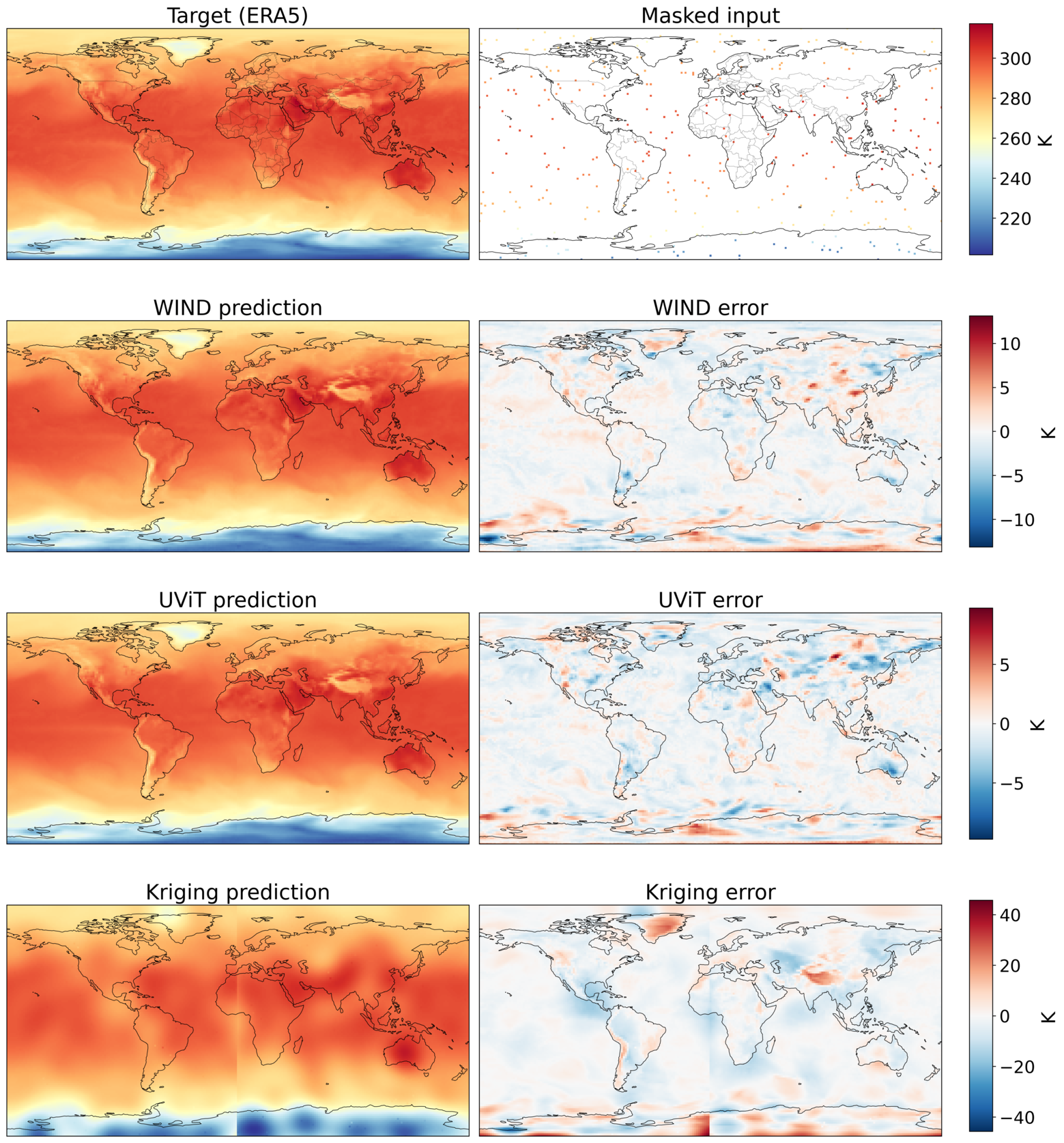}
    \caption{\textbf{Qualitative comparison of sparse reconstruction.} 
    \textbf{Left:} ERA5 ground truth (2m Temperature) and \textbf{1\%} sparse input mask. 
    \textbf{Center:} Predictions from \ourMethod, UViT, and Kriging. 
    \textbf{Right:} Prediction error relative to ground truth. 
    While \ourMethod and UViT recover physically coherent fields with realistic gradients, Kriging yields overly smooth interpolations that miss fine-grained weather patterns. \vspace{50em}}
    \label{fig:sparse_reconstruction_example_0_01}
\end{figure}

\begin{figure*}[t]
    \centering
    \includegraphics[width=0.95\linewidth]{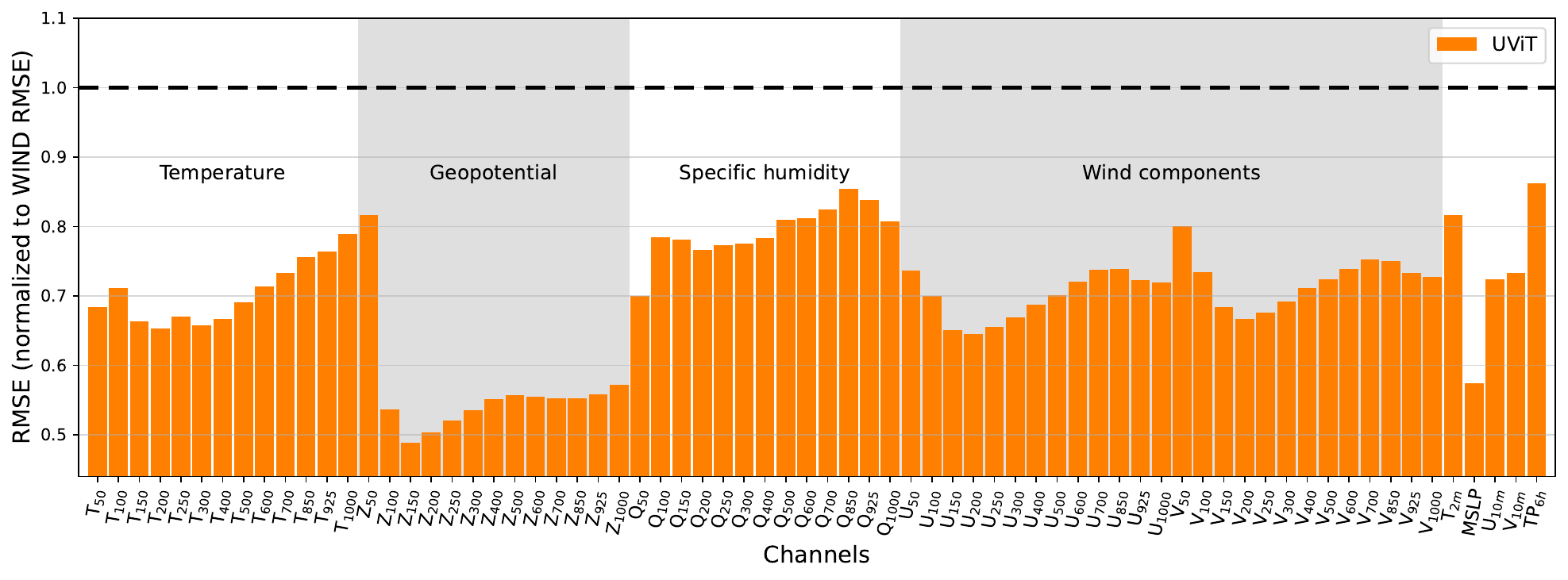}
    \caption{\textbf{RMSE comparison for sparse reconstruction (1\% sparsity).} We compare the RMSE of the specialized UViT baseline relative to \ourMethod (dashed line at 1.0). \ourMethod outperforms the specialized model (bars $>$ 1.0) on the majority of variables, particularly for fields like geopotential and specific humidity.}
    \label{fig:sparse_reconstruction_rmse_1_percent}
\end{figure*}

\begin{figure*}[b]
    \centering
    \includegraphics[width=0.95\linewidth]{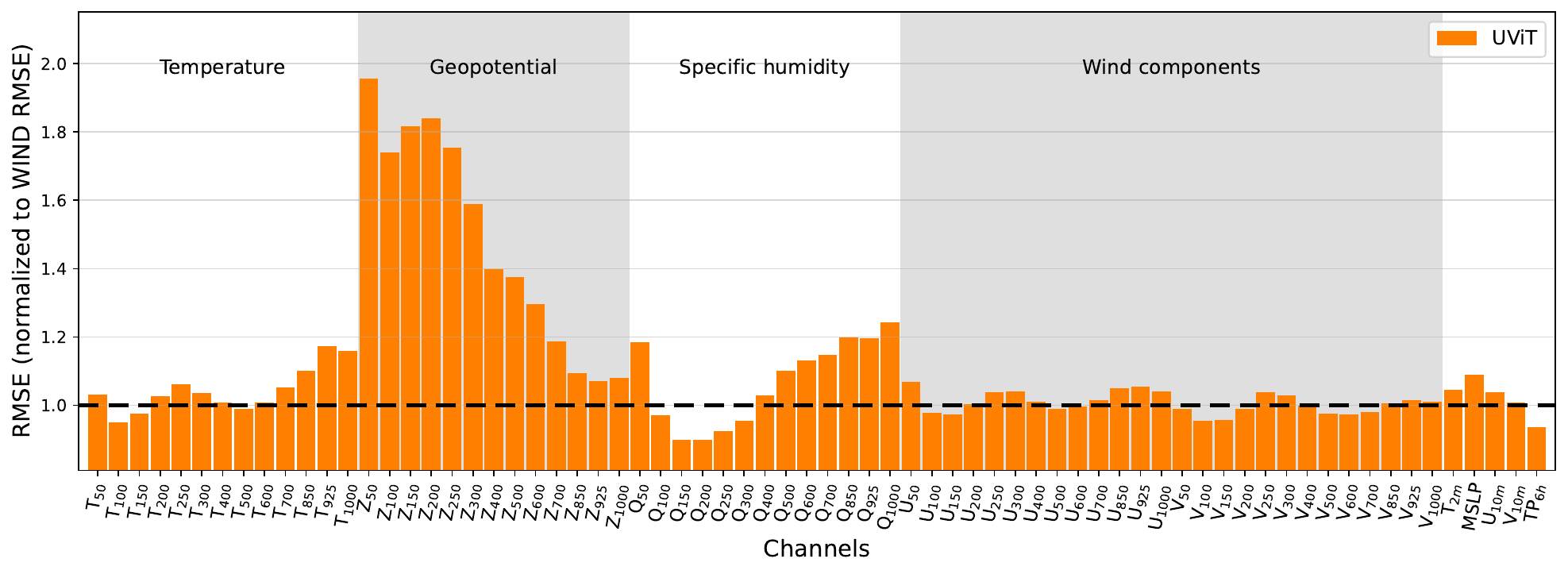}
    \caption{\textbf{RMSE comparison for sparse reconstruction (10\% sparsity).} We compare the RMSE of the specialized UViT baseline relative to \ourMethod (dashed line at 1.0). \ourMethod outperforms the specialized model (bars $>$ 1.0) on the majority of variables, particularly for fields like geopotential and specific humidity.}
    \label{fig:sparse_reconstruction_rmse_10_percent}
\end{figure*}

\begin{figure*}[t]
    \centering
    \includegraphics[width=0.99\linewidth]{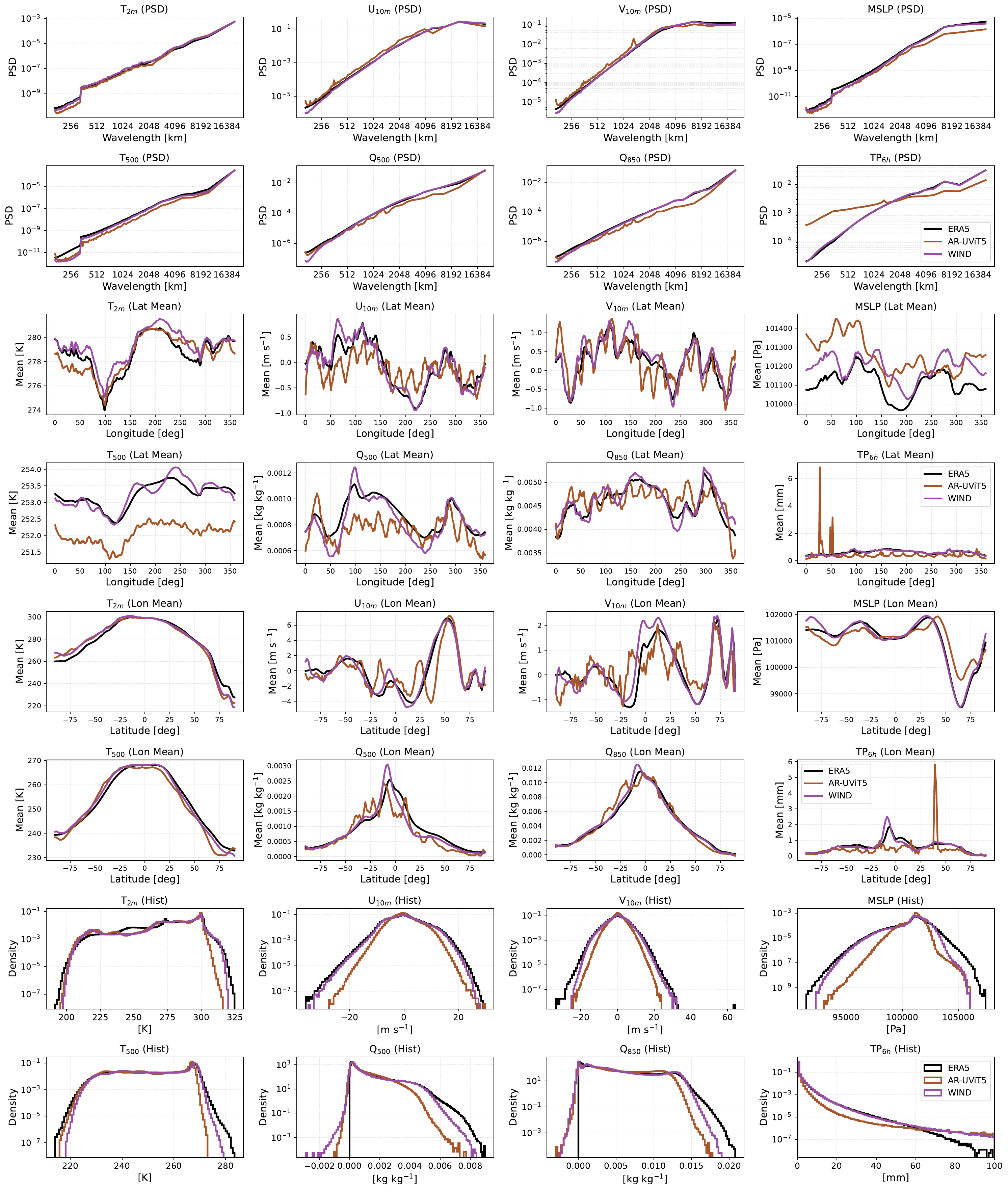}
     \caption{\textbf{Comparing rollout stability of diffusion forcing vs full sequence diffusion.} We compare the statistical properties of a 20-year unconstrained forecast generated by our model \ourMethod and the autoregressive baseline AR-UViT5 (full sequence diffusion) against the ground truth ERA5. Top rows: The PSD plots show that AR-UViT5 produces unphysical spikes across all variables, particularly for precipitation. In contrast, \ourMethod accurately preserves energy across the full spectrum. Middle rows: The latitudinal and longitudinal means show that \ourMethod is overall in sync with ERA5, while the baseline exhibits significant biases. Bottom rows: The histograms confirm that \ourMethod faithfully reproduces the full probability distributions only slightly overestimating the tails in precipitation. The baseline collapses completely for some distributions.
     } 
    \label{fig:20y_forecast_psd_latlon_mean_hist}
\end{figure*}

\begin{figure*}[t]
    \centering
    \includegraphics[width=0.99\linewidth]{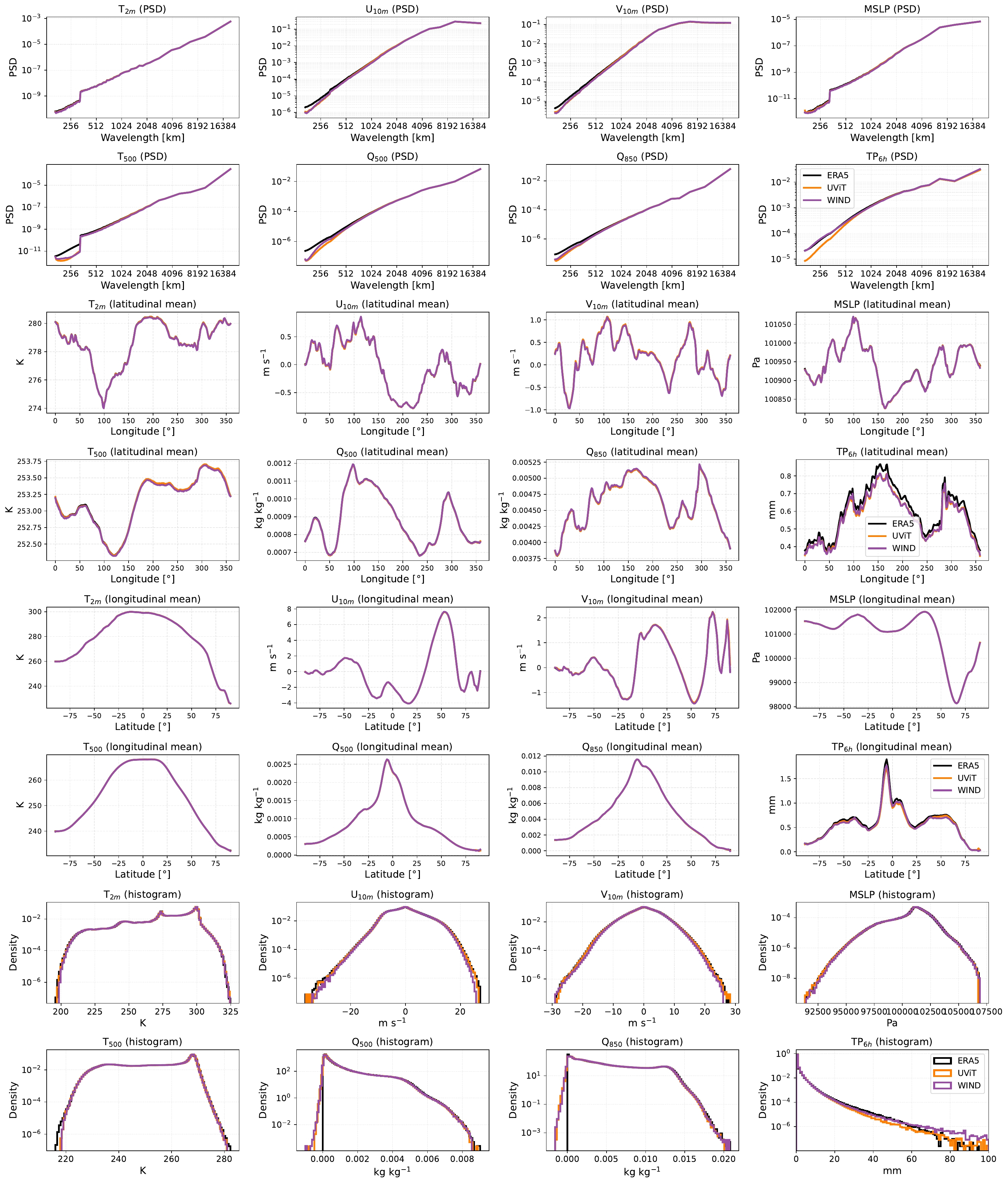}
     \caption{\textbf{Spectrum and distribution for temporally downscaled fields.} Top rows: \ourMethod agrees with frequency spectrum of ERA5 extremely well, even for precipitation where UViT struggles. Middle rows: The latitudinal and longitudinal means show that both models agree with ERA5. Bottom rows: The histograms confirm that both models reproduces the probability distributions of ERA5 well. \vspace{3em} }
    \label{fig:temporal_downscaling_psd_latlon_mean_hist}
\end{figure*}

\begin{figure*}
    \centering
    \includegraphics[width=\linewidth]{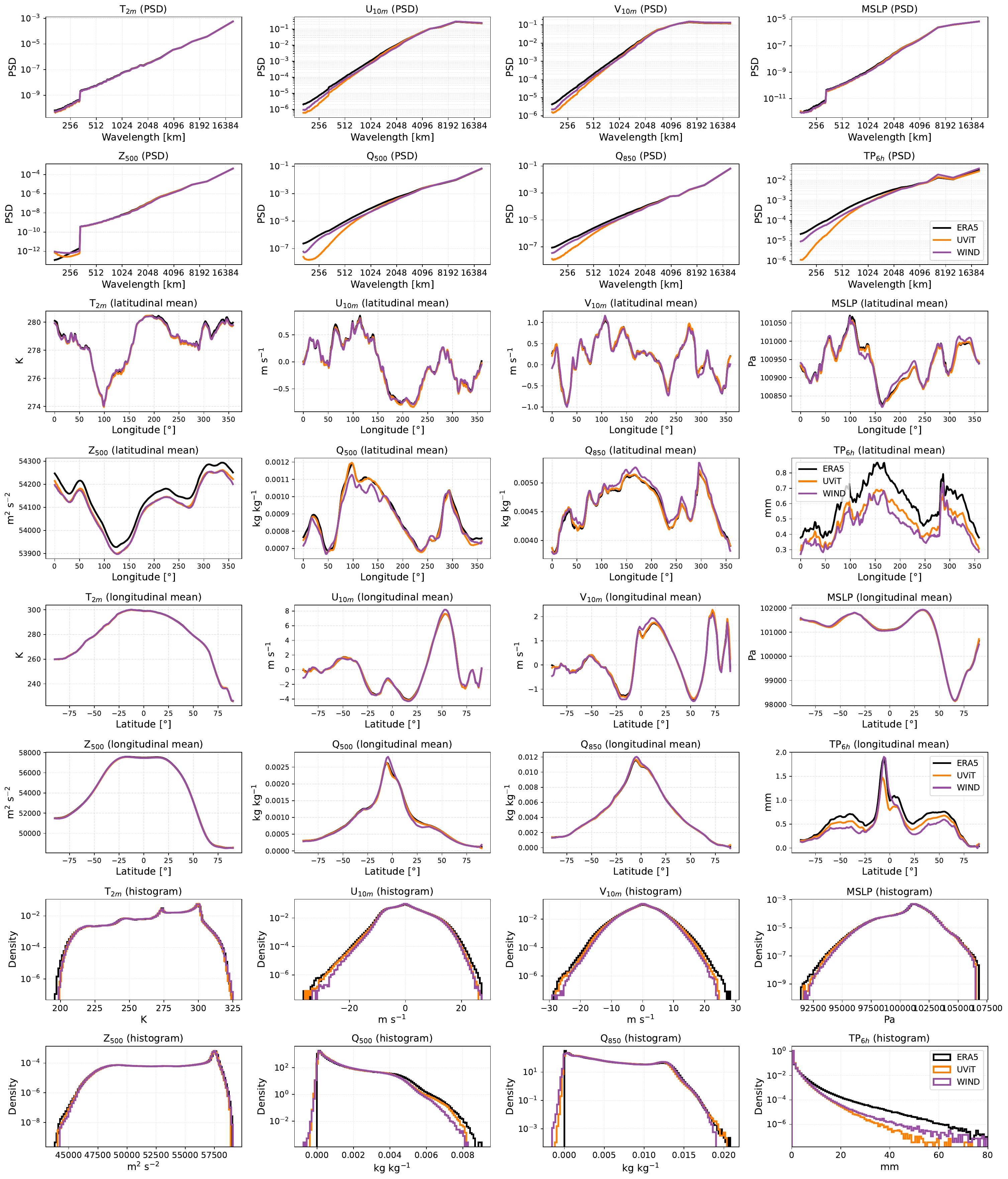}
    \caption{\textbf{Spectrum and distribution for sparse reconstruction (1 \% sparsity).} Top rows: \ourMethod agrees with frequency spectrum of ERA5, even at high frequencies where UViT struggles. Middle rows: The latitudinal and longitudinal means show that both models are in sync with ERA5, except for precipitation. Bottom rows: The histograms confirm that both models reproduces the full probability distributions. \ourMethod performs slightly better for precipitation.}
    \label{fig:sparse_reconstruction_psd_latlon_mean_hist_1_percent}
\end{figure*}

\begin{figure*}
    \centering
    \includegraphics[width=\linewidth]{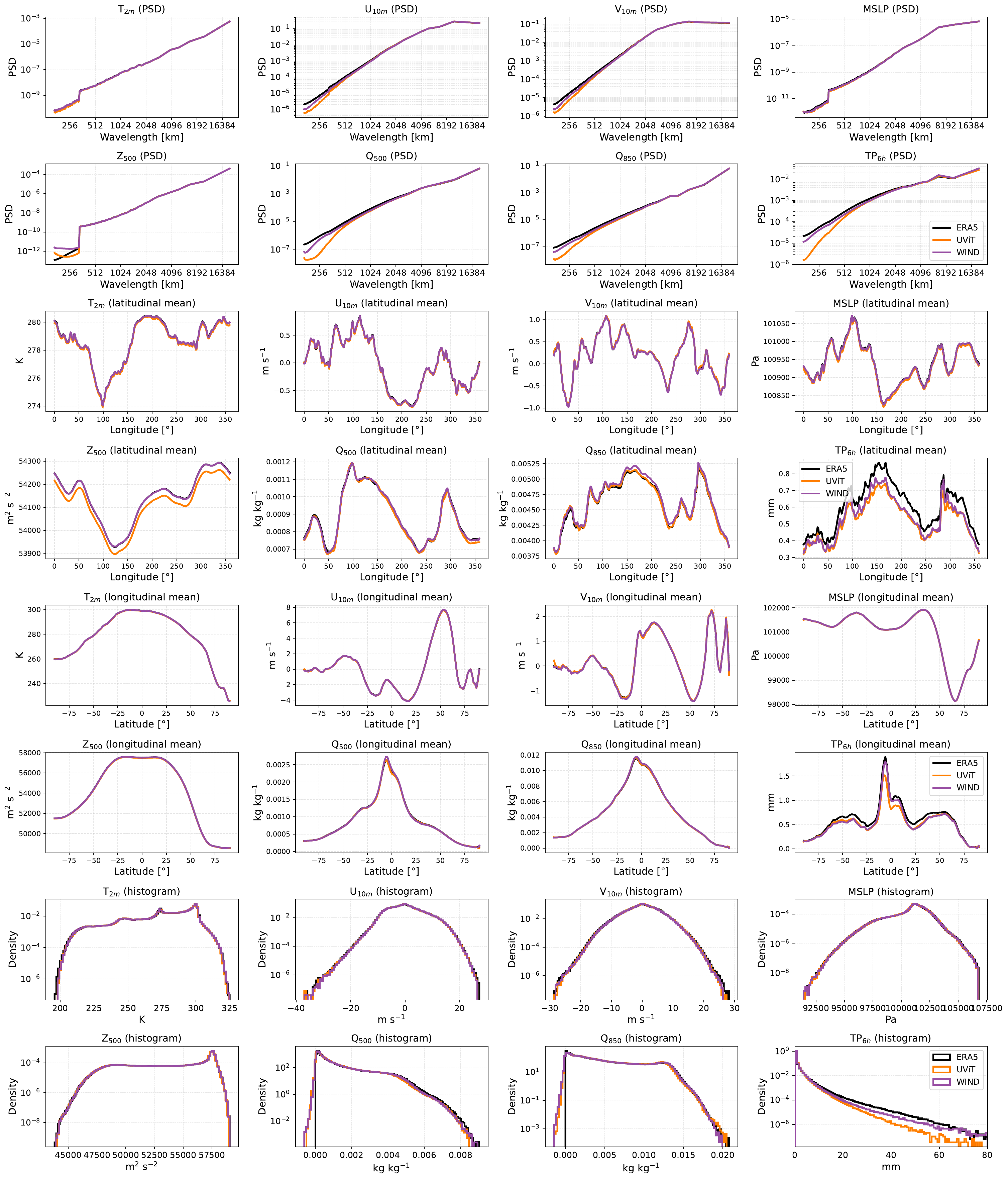}
    \caption{\textbf{Spectrum and distribution for sparse reconstruction (10 \% sparsity).} Top rows: \ourMethod agrees with frequency spectrum of ERA5, even at high frequencies where UViT struggles. Middle rows: The latitudinal and longitudinal means show that both models align well with ERA5, with slight deviations for precipitation. Bottom rows: The histograms confirm that both models reproduces the full probability distributions. \ourMethod is superior for approximating the histogram for precipitation. }
    \label{fig:sparse_reconstruction_psd_latlon_mean_hist_10_percent}
\end{figure*}

\end{document}